\documentclass{article}

\usepackage[main, final]{neurips_2025}

\usepackage[utf8]{inputenc} 
\usepackage[T1]{fontenc}    
\usepackage{hyperref}       
\usepackage{url}            
\usepackage{xurl}           
\usepackage{booktabs}       
\usepackage{amsfonts}       
\usepackage{nicefrac}       
\usepackage{microtype}      
\usepackage{xcolor}         
\usepackage[main, final]{neurips_2025}
\usepackage{makecell}
\usepackage{adjustbox}
\usepackage{amsmath} 
\usepackage{float}
\usepackage{appendix}
\usepackage{caption}
\usepackage{subcaption}

\title{Healthier LLMs: Retrieval-Augmented Generation for Public Health Question Answering}

\author{%
    Felix Feldman \And
    Joshua Harris \And
    Timothy Laurence \And 
    Leo Loman \And
    Ollie Higgins \And 
    Fan Grayson \And
    Poonam Soma \And
    Bethany Pace-Bonello \And
    Michael Borowitz \And
    Toby Nonnenmacher
}

\begin{document}

\maketitle

\begin{abstract}
Large language models (LLMs) achieve promising results on medical question answering benchmarks, yet their use in public health is constrained by hallucinations and the rapid evolution of official guidance. Retrieval-Augmented Generation (RAG) mitigates these risks by grounding responses in an explicitly maintained corpus, but end-to-end performance depends critically on retrieval configuration and on evaluation beyond multiple-choice formats. We extend PubHealthBench, a question answering (QA) benchmark of 7,929 questions derived from UK Government public health guidance, into a retrieval-augmented setting and systematically evaluate retrieval and generation choices. We compare dense, sparse, and hybrid retrieval across multiple embedding models and corpus variants, and show that hybrid retrieval consistently improves recall and ranking quality, with chunk length and topic interacting with ranking performance. Providing retrieved context substantially increases multiple-choice accuracy across a diverse set of LLMs, enabling smaller open-weight models to match or outperform larger  models used without retrieval, with gains primarily driven by retrieval quality and careful context selection. To assess realistic free-form answering, we introduce a rubric-based LLM-as-a-judge covering faithfulness, completeness, clarity, and factual consistency, and validate it against dual human annotations. Judge–human agreement is strongest for faithfulness and completeness, while factual consistency and clarity are less reliably reproduced, motivating caution when interpreting those dimensions at scale. Overall, our results highlight retrieval as a primary lever for reliable public health QA and provide practical guidance for building and evaluating RAG systems grounded in official guidance.
\end{abstract}

\section{Introduction}

Artificial intelligence (AI) is playing an expanding role in public health, from chatbots that answer health queries \cite{Harris2025PubHealthBench} to decision-support tools for public health professionals \cite{Editor_2025}. Large language models (LLMs) can already generate coherent answers based on extensive training data, in some cases approaching expert performance on medical question answering tasks \cite{singhal2022largelanguagemodelsencode}. This potential has spurred interest in deploying LLMs in public health, both as public-facing tools and as decision aids for public health professionals. Unlike many clinical decision-support settings—where tools are designed to support decisions for an individual patient at the point of care, public health guidance is population-level, often precautionary, and closely tied to official recommendations that are updated as evidence and policy evolve \citep{Sutton2020CDSS,ECDC2011EvidenceBasedPublicHealth,WHO2022LivingGuidelines}. However, LLMs can hallucinate information or provide outdated advice, which in public health can have severe implications: even small inaccuracies may harm individual health decision-making \cite{WHO2024DisinformationPublicHealth} and, at scale, drive inappropriate or dangerous behaviours across populations \cite{doNascimento2022InfodemicsReviewOfReviews}. Therefore, ensuring that LLM responses are reliable and up to date is a prerequisite for safe AI adoption in public health.

One established approach to improving LLM performance and reliability is \textbf{Retrieval-Augmented Generation} (RAG) \cite{lewis2020rag}. In RAG systems, an LLM is coupled with an external retrieval component that selects relevant documents from a knowledge base; the model then conditions its output on this retrieved context rather than relying solely on its parametric memory \cite{gao2024ragsurvey, sharma2025ragsurvey}. By incorporating retrieval, an LLM can expand and update its effective knowledge beyond what is stored in its frozen parameters, mitigating hallucinations and reducing the impact of outdated training data \cite{ni2025trustworthyragsurvey}. In high-stakes fields such as public health, the ability to provide answers grounded in trusted evidence is especially valuable. RAG systems have been shown to substantially improve performance on knowledge-intensive tasks such as medical and guideline-based question answering compared with LLMs that rely only on internal knowledge~\citep{Shi2024MKRAG, Vach2025NeuroRAG, fernandezpichel2025searchenginesllmshealth}.

With the increasing adoption of LLM systems in use cases involving public health or medical queries, it is crucial to measure their effectiveness on tasks that reflect real information needs of both the public and professionals \cite{Yan2024MedLLM-Benchmarks, ventura2025healthqa-br, arora2025healthbenchevaluatinglargelanguage}. To this end, a variety of benchmarks for medical Question Answering (QA) have been developed, though most existing benchmarks focus on clinical or biomedical questions rather than public health guidance. To understand this issue for UK Government public health information Harris et al. evaluated LLMs on a range of public health classification and extraction tasks \cite{harris2025evaluatinglargelanguagemodels} before introducing \textbf{PubHealthBench} \cite{Harris2025PubHealthBench}, a QA  benchmark with nearly $8{,}000$ questions derived from official UK public health guidance.

In this follow-up work, we explore RAG as a solution to improve LLM performance on public health QA. We extend our benchmark setup by allowing models to retrieve information from the same underlying corpus of UK public health guidance from which the questions were generated~\citep{Harris2025PubHealthBench}. Specifically, we use PubHealthBench to address four questions. (1) How effective are different retrieval setups, spanning embedding models and system design choices, at identifying the most relevant chunks of public health guidance given questions about that guidance? (2) Does introducing retrieval of trusted documents improve LLM performance on multiple-choice QA (MCQA), and how does this depend on retrieval quality and context configuration? (3) To what extent do the observed MCQA improvements generalise to free-from responses? (4) Can we develop an automated evaluation framework to assess LLM responses that aligns with human expert decisions? By answering these questions, we aim to demonstrate how coupling LLMs with a corpus of public health guidance can yield more reliable and informed responses on knowledge-intensive tasks.

\section{Related Work}
\subsection{Retrieval-Augmented Generation Foundations}
RAG combines external knowledge retrieval with large language model (LLM) generation to enhance factual accuracy and mitigate hallucinations \cite{lewis2020rag, Shi2024MKRAG}. For example, the RETRO model demonstrated that pairing a smaller generator with a large retrieval data store can match much larger LLM-only systems \cite{Borgeaud2022RETRO}, and subsequent work has shown that retrieval scale and quality are critical performance levers \cite{Shao2024DatastoreScaling}. Recent surveys provide up-to-date taxonomies of RAG architectures \cite{Cheng2025KnowledgeOrientedRAGSurvey, gao2024ragsurvey, sharma2025ragsurvey} and examine trust, safety, fairness, and accountability in retrieval-augmented systems \cite{ni2025trustworthyragsurvey}. Together, these works highlight retrieval as a distinct driver of performance and underscore the challenge of reliably integrating retrieved knowledge, particularly in safety-critical domains such as medicine and public health.

\subsection{Retrieval Methods and Context Handling}
Modern retrieval systems largely rely on dense semantic embeddings to map text into vector space representations, with trade-offs between model size, retrieval accuracy, and latency \cite{fang2024scalinglawsdenseretrieval}. Hybrid retrieval methods that combine dense embeddings with sparse models such as BM25 often boost retrieval precision and recall \cite{Bruch2023AnalysisFusionHybridRetrieval}. Re-ranking models (e.g., GTE-ModernColBERT) and knowledge-graph-based retrieval further support retrieving relevant information in domain-specific settings \cite{GTE-ModernColBERT, wu2024medicalgraphragsafe, Boeckling2024WalkRetrieve}. Importantly, long context windows change but do not remove the need for careful retrieval and context selection \cite{Pratapa2025EstimatingContext, jin2025longcontext}. There is also extensive empirical work evaluating retrieval design choices, such as, query expansion, chunking, memory-based architectures; underlining how there are several configuration choices that can impact retrieval, and downstream QA, performance \cite{Li2025EnhancingRAGBestPractices, Qin2025AdaptiveMemoryRAG}.

\subsection{Benchmarking and Evaluation in Health QA}

In medical and public-health domains, multiple-choice QA benchmarks (e.g., MedMCQA, MedQA) have provided standard evaluation frameworks for assessing clinical knowledge \cite{pmlr-v174-pal22a, Jin2021WhatDD, Yan2024MedLLM-Benchmarks}. However, multiple-choice formats can overestimate model competence compared with free-form generation tasks and may mask important reasoning failures \cite{singh2025mcq}. For example, Singh et al.\ report an average drop in accuracy of 39.4\% when moving from MCQA to free-from answers for LLMs (vs.\ 22.3\% for humans). They also show that even with the question stem fully masked, MCQA accuracy remains above chance (6.7\% above random on average), suggesting MCQA performance can be partly driven by answer-option cues rather than underlying understanding \cite{singh2025mcq}. Consequently, newer benchmarks have shifted toward free-form responses to better simulate real-world usage \cite{hosseini2024medicalQA, manes2023kqa}. In public health, studies show substantial knowledge gaps and uneven performance across domains: AfriMed-QA reports that GPT-4o attains $79\%$ overall accuracy on expert MCQs, but performance varies substantially by specialty—ranging from $>90\%$ in top specialties (e.g., rheumatology) to $<60\%$ in weaker areas such as pediatrics and obstetrics–gynecology \cite{nimo-etal-2025-afrimed}, illustrating strong topic dependence despite high aggregate QA accuracy, and PubHealthBench finds that while the strongest models achieve $>90\%$ accuracy ($92.5\%$ for GPT-4.5) in MCQA, exceeding the $88\%$ human baseline, but no model scored higher than 75\% in free-form QA\cite{Harris2025PubHealthBench}. 

Evaluation methods are also evolving. Overlap metrics such as ROUGE or BLEU correlate poorly with human judgments in free-form QA \cite{xian2025lfqa, xu2023lfqa-eval}. The use of LLMs as automated evaluators (“LLM-as-Judge”) has emerged as a scalable alternative, with specialised models such as Prometheus and its successor showing strong alignment with human assessments across a range of generation tasks \cite{Kim2023Prometheus, Kim2024Prometheus2, Zheng2023MTBench, Gu2024LLMJudgeSurvey}. In health-specific settings, rubric-guided LLM-as-judge frameworks that combine physician-designed rubrics with model graders have achieved high agreement with clinician ratings \cite{arora2025healthbenchevaluatinglargelanguage, Croxford2025MedLLMJudge}. A recent comparative study of search engines, LLMs, and RAG variants on health questions further demonstrates retrieval’s substantial impact on QA accuracy and highlights the benefits of grounding answers in retrieved evidence \cite{fernandezpichel2025searchenginesllmshealth}. Together, these developments underscore the need for benchmarks that jointly assess retrieval quality and generation performance in health QA systems, especially when systems are expected to align with official public health guidance.

\section{Methods}
\label{sec:methods}

\subsection{Benchmark and Dataset}
We use the PubHealthBench QA benchmark, which contains $7,929$ multiple‐choice questions derived from 687 UK Government public health guidance documents covering 10 topic areas \cite{Harris2025PubHealthBench}. The source documents were converted to markdown and split into chunks by markdown header levels, with full header hierarchy appended for context. This produced $5,358$ chunks, which form the retrieval corpus.
\begin{table}[H]
  \caption{PubHealthBench subsets \citep{Harris2025PubHealthBench}.}
  \label{tab:pubhealthbench_subsets}
  \centering
  \begin{adjustbox}{max width=\linewidth}
    \begin{tabular}{@{}l p{0.07\linewidth} p{0.45\linewidth} p{0.12\linewidth} p{0.35\linewidth}@{}}
      \toprule
      Subset & Size & Creation method & QA format & Purpose \\
      \midrule
      \textbf{PubHealthBench-Full}
        & 7{,}929
        & LLM generated multiple-choice questions created from single chunks of public health guidance via an automated pipeline.
        & MCQA
        & {\raggedright Broad coverage to assess LLM performance across many public health topics and guidance audiences.\par} \\
      \textbf{PubHealthBench-FreeForm}
        & 760
        & Random subset of questions manually reviewed by experts (with ambiguous/invalid items identified), presented without multiple-choice options.
        & Free-form
        & {\raggedright More realistic open-ended evaluation of LLM performance on free-form QA.\par} \\
      \bottomrule
    \end{tabular}
  \end{adjustbox}
\end{table}

\subsection{Retrieval}
We adopt five retrieval methods for our RAG pipeline:
\begin{enumerate}
  \item \textbf{Embedding‐based retrieval:} Each chunk is encoded into a dense embedding via a text‐embedding model, and queries are encoded with the same model. Retrieval ranks chunks by cosine similarity between query and chunk embeddings. We evaluate eight embedding models (e.g., NV-Embed-V2 \cite{lee2024nvembed}, EmbeddingGemma \cite{google2025embeddinggemma_blog}, ModernBertBase \cite{modernbert2024}, SFR-Embedding-Mistral \cite{meng2024sfr_embedding_mistral}, Multilingual-E5-large \cite{wang2024multilingualE5}, OpenAI’s text-embedding-3-large \cite{openai2024textembedding3large}).
  \item \textbf{Keyword-based retrieval:} We build sparse indices using the term frequency algorithms TF-IDF and BM25. Retrieval ranks chunks by the overlap of key terms between the query and chunks.
  \item \textbf{Hybrid retrieval:} We perform both embedding-based and keyword-based retrieval as in (1) and (2), and then merge the two ranked lists using a weighted Reciprocal Rank Fusion (RRF):
    \begin{equation}
      R(d;\alpha) = \alpha\,\frac{1}{c + r_{\mathrm{dense}}(d)} + (1-\alpha)\,\frac{1}{c + r_{\mathrm{sparse}}(d)}
      \label{eq:rrf}
    \end{equation}
    Where $\alpha$ is a weighting factor, we set \(c = 60\) consistent with the value used in foundational RRF work \cite{Cormack2009ReciprocalRankFusion}.
  \item \textbf{Summary‐based hybrid retrieval:} Each chunk is summarised using GPT-4o‐mini \cite{dangi2025evaluationgpt4ogpt4ominisvision}. Hybrid retrieval (as in (3)) is applied to the summary corpus rather than full chunks. This variant tests whether more compact context improves retrieval precision in the public-health domain. Summaries are used only for retrieval and the corresponding full chunk is passed down-stream for generation. 
  \item  \textbf{Reduced-corpus hybrid retrieval:} Hybrid retrieval (as in (3)) is applied to a reduced corpus, where any chunk longer than 512 tokens (for a given embedding model's tokenizer) is replaced with its generated summary (the 512-token cutoff aligns with the context capacity of Multilingual-E5 \cite{wang2024multilingualE5}). As with (4), only the corresponding full chunk is passed down-stream for generation.
\end{enumerate}

\subsection{Generation}
The top‐\(k\) ranked chunks from retrieval are appended (with a chunk identifier and separator) and concatenated into a context block. Following the prompt structure of Harris et al. \cite{Harris2025PubHealthBench}, we insert the question and the context block, with a simple instruction to make use of the inserted context when answering. We assess 11 LLMs (e.g., Llama 3.3 \cite{grattafiori2024llama3herd}, Phi-4 \cite{abdin2024phi4}, Gemma-3 \cite{gemma32025}, MedGemma \cite{sellergren2025medgemma}, CommandR \cite{cohere2024command_r_v01}) on the benchmark subsets shown in Table \ref{tab:pubhealthbench_subsets}.

\subsection{Experimental Setup}
\label{sec:exp_setup}

\subsubsection{Retrieval Setup}
Retrieval is run with each embedding model over all $7,929$ queries on three corpora:  
\(\mathcal{C}_F\) (full corpus), \(\mathcal{C}_S\) (summary-only corpus), and \(\mathcal{C}_R\) (reduced corpus). For model and each hybrid retrieval configuration we find the the value of the weighting factor $\alpha \in \{0.50, 0.55, 0.6, ..., 0.95\}$ that results in the best retrieval performance, where higher $\alpha$ means heavier weighting for embedding-based ranking (see Equation \ref{eq:rrf}). We report the optimal retrieval setup (method, corpus, $\alpha$) for each model. 

\subsubsection{MCQA Setup}
 For \textbf{PubHealthBench-Full}, we extract top-\(k\) chunks with \(k \in \{1,3,5,10\}\) from three representative retrieval configurations (NV-Embed-v2 ($\mathcal{C}_R$), Multilingual-E5-large-instruct ($\mathcal{C}_R$), ModernBertBase ($\mathcal{C}_F$)). These models are chosen to test the impact of retrieval quality on MCQA accuracy, as they span a range of model sizes and retrieval performance. We report MCQA accuracy on \textbf{PubHealthBench-Full}, alongside accuracy without retrieval of the human baseline and GPT-4.5 \cite{Harris2025PubHealthBench}.
 
 \subsubsection{Free-form Setup}
 In a large real world guidance corpus to achieve near perfect ($\sim 100\%$) retrieval recall, we may require a large number of retrieved chunks, so for free form responses we use the same three configurations for \textbf{PubHealthBench-FreeForm}, set \(k=15\), and limit the context block to $10,000$
 tokens. Through this setup we evaluate LLMs in a more realistic setting, where they are challenged to precisely extract relevant information from a large amount of context, and use this to clearly and accurately answer a public health query, without guiding answer options. In this setup, we evaluate free-form answers using an LLM judge (see \S ~\ref{sec:evaluation}).

\subsection{Evaluation}
\label{sec:evaluation}
\subsubsection{Retrieval Evaluation}
Because each query has exactly one target chunk, we report the following metrics:  
\begin{itemize}
  \item \textbf{Recall@\(k\)}: Proportion of queries whose relevant chunk appears within the top \(k\). Reflects coverage of retrieval but all rankings in the top $k$ are treated equally. 
  \item \textbf{Mean Reciprocal Rank (MRR)}: The average of \(1/r\) across queries, where \(r\) is the rank of the relevant chunk. Measures how early the correct chunk appears in the ranking.
  \item \textbf{Normalized Discounted Cumulative Gain (nDCG@\(k\))}: Assesses ranking quality via log-discounted gain, penalising relevant chunks appearing lower in the ranking \cite{thakur2021beir, jeunen2024ndcg}. This provides a finer-grained view of rank quality.
  \item \textbf{Precision@1}: Because there is exactly one relevant chunk per query, we report Precision@1 (1 if the chunk is ranked first, else 0) \cite{wang2024birco}.  
\end{itemize}
Together, these metrics allow us to evaluate whether the correct chunk is retrieved at all (Recall@$k$), how early it appears (MRR), and how well the ranking order is structured (nDCG).
\subsubsection{Free-form Evaluation}
 To evaluate the answers for \textbf{PubHealthBench-FreeForm}, we follow HealthBench in applying a rubric-based LLM-as-a-Judge to evaluate free form responses \cite{arora2025healthbenchevaluatinglargelanguage}. Through this we aim to create a framework for automated evaluation of free-form responses to public health queries. In our setup, each answer is assessed by the GPT-OSS-120B model, using a structured set of four criteria on factual consistency, completeness, clarity, and faithfulness to official guidance. The criteria are designed to be non-overlapping and objective, assessing an LLMs ability select only relevant information from a large amount of context, and use this to accurately and clearly answer a public health query. The criteria definitions are shown in in Table \ref{tab:criteria_defs}. To support reliable analysis, two human expert reviewers independently annotated the same sample of 100 LLM responses against each criterion. We report Cohen’s $\kappa$ and macro-F1, with 95\% bootstrap confidence intervals for the agreement between the reviewers and each reviewer with the LLM judge. All judgments are made with respect to the single relevant chunk of guidance, along with a reference answer, question and LLM answer, and detailed instructions for each criteria, aligning with HealthBench’s emphasis on grounded, context-aware evaluation \cite{arora2025healthbenchevaluatinglargelanguage}.
\begin{table}[H]
\centering
\caption{LLM-as-a-Judge criteria definitions used to evaluate free-form responses.}
\tiny
\begin{tabular}{p{3cm}p{10cm}}
\toprule
\textbf{Criterion} & \textbf{Definition} \\
\midrule
\textbf{Faithfulness} & The answer must not introduce any new claims, advice, or information not found in the provided official guidance. \\ 
\textbf{Completeness} & The answer should include all key points and recommendations from the guidance necessary to fully address the question. \\ 
\textbf{Factual Consistency} & The parts of the answer that relate to the guidance must be factually accurate and reflect the intended meaning. \\ 
\textbf{Clarity} & The answer should clearly communicate its main point and be grammatically correct, focused, and easy to follow. \\
\bottomrule
\end{tabular}
\label{tab:criteria_defs}
\end{table}

\section{Results}
\subsection{Retrieval}
\label{sec:ret_results}
We find that the retrieval performance of embedding models varies considerably on the task of retrieving relevant public health information. In dense-only retrieval configuration, the best performing model, \textbf{NV-Embed-v2}, achieves the best results across all metrics ($0.98$ Recall@10, $0.85$ MRR, $0.88$ nDCG@10), a marked gain over other models tested; for example, \textbf{EmbeddingGemma} scores $49$ ppts lower on Precision@1. Retrieval quality generally improves with model size and embedding dimension, yet sparse-only retrieval using TF-IDF or BM25 outperforms some embedding models. For every model, a hybrid setup improves performance across all retrieval metrics; the optimal fusion weight $\alpha$ in our RRF setup is model-dependent. Notably, hybrid search narrows the gap between smaller encoders (\textbf{multilingual-E5-large}, \textbf{ModernBERT-base}), which see a $5$--$10$ ppt increase across metrics, and \textbf{NV-Embed-v2}, despite a factor of $10$ difference in parameter count. No model attains its best results on the summary-only corpus $(\mathcal{C}_S)$; peak scores arise on either the reduced $(\mathcal{C}_R)$ or full $(\mathcal{C}_F)$ corpora, and this preference is not explained by context length alone.
\begin{table}[H]
\centering
\caption{\textbf{Retrieval metrics for sparse and dense-only retrieval} Metrics across embedding models and sparse baselines. NV-Embed-v2 is the strongest model in a dense-only setup.}
\label{tab:model_only_retrieval}
\scriptsize
\begin{adjustbox}{max width=\linewidth}
\begin{tabular}{lrrrccccc}
\toprule
& \multicolumn{3}{c}{\textbf{Model metadata}} & \multicolumn{5}{c}{\textbf{Retrieval metrics}} \\
\cmidrule(lr){2-4}\cmidrule(lr){5-9}
\textbf{Embedding model} & \textbf{Params} & \textbf{Dim} & \textbf{Context len.} & \textbf{Precision@1} & \textbf{Recall@5} & \textbf{Recall@10} & \textbf{MRR} & \textbf{nDCG@10} \\
\midrule
NV-Embed-v2                    & 8B   & 4096 & 32768 & \textbf{0.76} & \textbf{0.96} & \textbf{0.98} & \textbf{0.85} & \textbf{0.88} \\
SFR-Embedding-Mistral          & 7B   & 4096 & 32768 & 0.71 & 0.94 & 0.97 & 0.81 & 0.85 \\
text-embedding-3-large         & --   & 3072 & 8192  & 0.68 & 0.92 & 0.96 & 0.78 & 0.83 \\
Multilingual-e5-large          & 0.6B & 1024 & 512   & 0.68 & 0.91 & 0.96 & 0.78 & 0.83 \\
Multilingual-e5-large-instruct & 0.6B & 1024 & 512   & 0.64 & 0.90 & 0.95 & 0.76 & 0.80 \\
EmbeddingGemma                 & 0.3B & 768  & 2048  & 0.27 & 0.55 & 0.69 & 0.41 & 0.47 \\
ModernBERT-base                & 0.15B & 768  & 8192  & 0.63 & 0.90 & 0.95 & 0.75 & 0.79 \\
MedCPT-Query-Encoder           & 0.1B & 768  & 512   & 0.24 & 0.51 & 0.63 & 0.37 & 0.42 \\
TF-IDF                         & --   & --   & --    & 0.61 & 0.86 & 0.92 & 0.72 & 0.77 \\
BM25                           & --   & --   & --    & 0.65 & 0.87 & 0.91 & 0.75 & 0.79 \\
\bottomrule
\end{tabular}
\end{adjustbox}
\end{table}
\begin{table}[H]
\centering
    \caption{\textbf{Best-performing retrieval configurations} Retrieval metrics for models in hybrid retrieval setups. NV-Embed-v2 remains the strongest but other models show significant improvements in hybrid setups.}
\label{tab:best_retrieval}
\scriptsize
\begin{adjustbox}{max width=\linewidth}
\begin{tabular}{lccccccccc}
\toprule
& \multicolumn{4}{c}{\textbf{Retrieval configuration}} & \multicolumn{5}{c}{\textbf{Retrieval metrics}} \\
\cmidrule(lr){2-5}\cmidrule(lr){6-10}
\textbf{Embedding model} 
& \textbf{Corpus} 
& \textbf{Method} 
& \textbf{Sparse index} 
& $\boldsymbol{\alpha}$ 
& \textbf{Precision@1} 
& \textbf{Recall@5} 
& \textbf{Recall@10} 
& \textbf{MRR} 
& \textbf{nDCG@10} \\
\midrule
NV-Embed-v2                    
& $\mathcal{C}_R$ & hybrid & TF-IDF & 0.75 
& \textbf{0.80} & \textbf{0.97} & \textbf{0.99} & \textbf{0.88} & \textbf{0.91} \\

SFR-Embedding-Mistral          
& $\mathcal{C}_F$ & hybrid & TF-IDF & 0.75 
& 0.76 & 0.96 & 0.99 & 0.85 & 0.88 \\

text-embedding-3-large         
& $\mathcal{C}_F$ & hybrid & TF-IDF & 0.55 
& 0.77 & 0.96 & 0.98 & 0.85 & 0.88 \\

Multilingual-e5-large          
& $\mathcal{C}_R$ & hybrid & TF-IDF & 0.80 
& 0.76 & 0.95 & 0.98 & 0.84 & 0.87 \\

Multilingual-e5-large-instruct 
& $\mathcal{C}_R$ & hybrid & TF-IDF & 0.75 
& 0.75 & 0.95 & 0.98 & 0.84 & 0.87 \\

EmbeddingGemma                 
& $\mathcal{C}_R$ & hybrid & TF-IDF & 0.50 
& 0.63 & 0.88 & 0.93 & 0.74 & 0.78 \\

ModernBERT-base                
& $\mathcal{C}_F$ & hybrid & TF-IDF & 0.55 
& 0.74 & 0.94 & 0.98 & 0.83 & 0.86 \\

MedCPT-Query-Encoder           
& $\mathcal{C}_F$ & hybrid & TF-IDF & 0.50 
& 0.64 & 0.88 & 0.94 & 0.74 & 0.79 \\

\bottomrule
\end{tabular}
\end{adjustbox}
\end{table}

\begin{figure}[t]
    \centering
    \includegraphics[width=\linewidth]{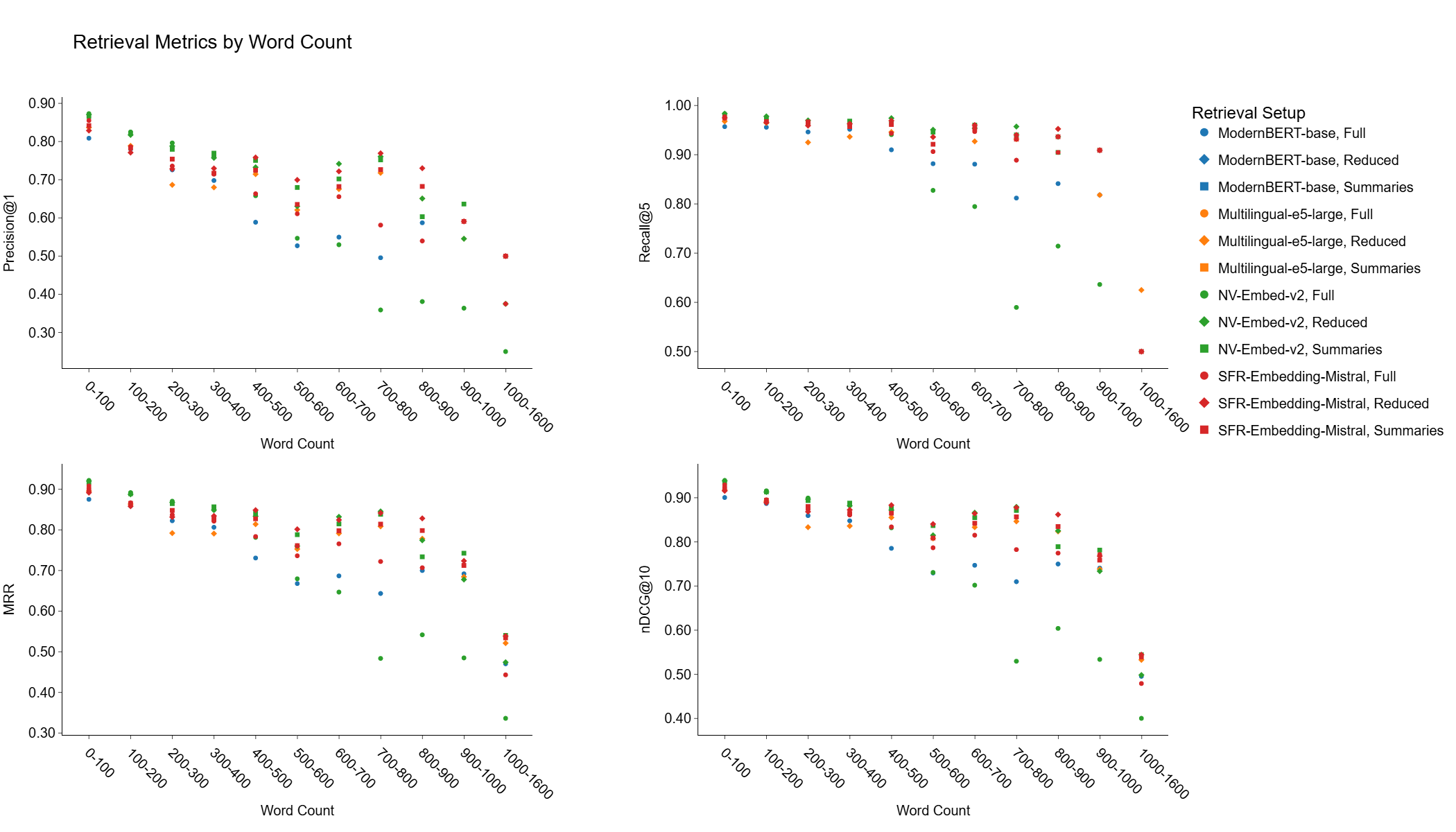}
    \caption{\textbf{Retrieval by target length} Retrieval metrics for various setups broken down by word count of the target chunk.}
    \label{fig:retrieval by len}
\end{figure}

Figure \ref{fig:retrieval by len} shows retrieval performance declines across models as target chunk length increases. Recall decreases only modestly, but rank-sensitive metrics (Precision@$k$, MRR, nDCG@$k$) drop sharply once chunks exceed $700$--$800$ words. Using $\left(\mathcal{C}_R\right)$ somewhat attenuates this decline relative to $\left(\mathcal{C}_F\right)$, although it does not mitigate it completely. As target length grows, the performance gap between larger models (\textbf{NV-Embed-v2}, \textbf{SFR-Embedding-Mistral}) and smaller models widens.

\begin{figure}[t]
    \centering
    \includegraphics[width=\linewidth]{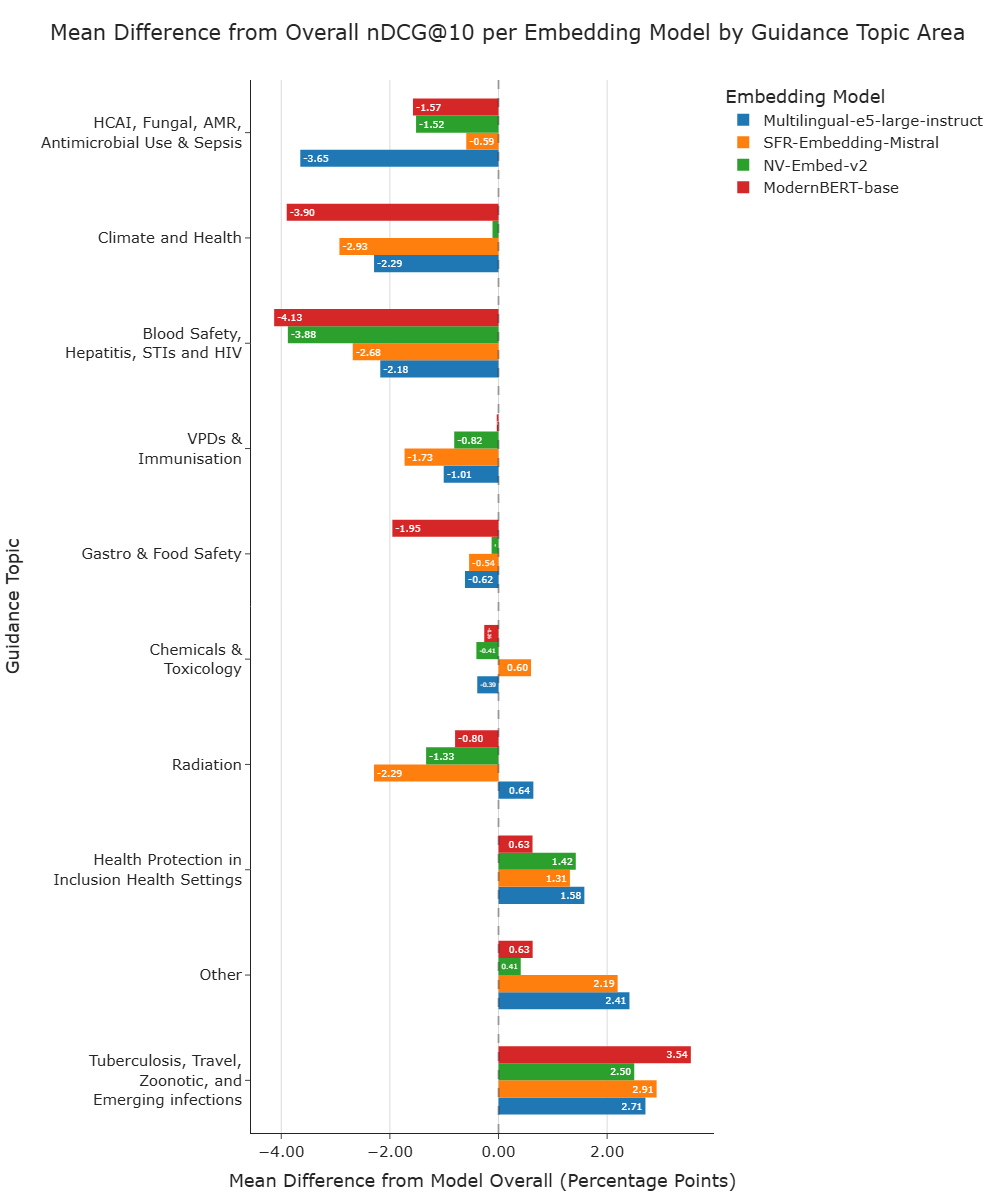}
    \caption{\textbf{Topic-wise deviations in ranking quality.} Bars show the mean difference (percentage points) between each guidance topic area's nDCG@10 and the corresponding model's overall mean nDCG@10 (dashed line at $0$)}
    \label{fig:retrival_avg_diff}
\end{figure}

In Figure \ref{fig:retrival_avg_diff}, we observe topic-level deviations in ranking quality across embedding models, measured as each topic's departure from the model's mean nDCG@10. A one-way ANOVA indicates that guidance topic explains less than 1\% of the variance in nDCG@10 across models ($\eta^{2} \approx 0.006$--$0.008$). Although this global effect size is small, topic means show systematic differences in ranking quality. The gap between lower-performing topics (e.g., \emph{Blood Safety, Hepatitis, STIs and HIV}) and the highest-performing topic (\emph{Tuberculosis, Travel, Zoonotic and Emerging Infections}) is approximately $0.06$--$0.08$ nDCG@10. The relative ordering of higher- and lower-performing topics is broadly consistent across models.

In contrast, Recall@5 shows little evidence of topic dependence ($\eta^{2} \approx 0.002$--$0.005$). Topic therefore has minimal influence on whether relevant content appears within the top-$k$ results, but it does affect the position at which relevant chunks are ranked once retrieved.

\subsection{MCQA with Retrieval}
Giving LLMs access to retrieved context significantly improves their MCQA performance; topic-wise accuracy per model is shown in Table \ref{tab:best_mcqa}. Other than \textbf{Gemma-3-1B}, all models exceed the human baseline (with cursory search engine use) of $88\%$ \cite{Harris2025PubHealthBench}. Several models also meet or surpass the best performing model without retrieved context (GPT\mbox{-}4.5, $92.5\%$) \cite{Harris2025PubHealthBench}. To summarise topic-wise stability, we report the coefficient of variation $\mathrm{CV}=\sigma/\mu$ across topic accuracies. With the exception of \textbf{Gemma-3-1B}, all models have $\mathrm{CV}<0.05$, indicating low between\mbox{-}topic variability compared with LLM\mbox{-}only results \cite{Harris2025PubHealthBench}.
\begin{table}[t]
\caption{\textbf{MCQA accuracy by guidance topic.} Best MCQA results per generation model across PubHealthBench guidance topic areas.}
\label{tab:best_mcqa}
\centering
\scriptsize
\begin{adjustbox}{max width=\textwidth}
\begin{tabular}{lrrrrrrrrrrr}
\toprule
\textbf{Generation model}
& \makecell{\textbf{Blood safety}\\\textbf{\& STIs/HIV}}
& \makecell{\textbf{Chemicals}\\\textbf{\& tox}}
& \makecell{\textbf{Climate}\\\textbf{\& health}}
& \makecell{\textbf{Gastro}\\\textbf{\& food}}
& \makecell{\textbf{HCAI/AMR}\\\textbf{\& sepsis}}
& \makecell{\textbf{Inclusion}\\\textbf{health}}
& \textbf{Other}
& \textbf{Radiation}
& \makecell{\textbf{TB/travel}\\\textbf{\& zoonoses}}
& \makecell{\textbf{VPDs}\\\textbf{\& imm.}}
& \textbf{Total} \\
\midrule
Llama-3.3-70B$^{\ast}$
                  & \textbf{0.995} & \textbf{0.997} & \textbf{1.000} & \textbf{0.994} & \textbf{1.000} & \textbf{0.998} & 0.992 & \textbf{1.000} & \textbf{0.992} & \textbf{0.997} & \textbf{0.995} \\
Command-R-32B     & 0.989 & 0.988 & 0.997 & 0.979 & 0.992 & 0.992 & 0.983 & 0.974 & 0.986 & 0.991 & 0.987 \\
MedGemma-27B      & 0.995 & 0.995 & \textbf{1.000} & 0.988 & 0.992 & 0.996 & \textbf{0.996} & 0.983 & 0.990 & 0.993 & 0.992 \\
Gemma-3-27B       & 0.992 & 0.992 & \textbf{1.000} & 0.988 & 0.992 & 0.996 & 0.992 & 0.987 & 0.990 & 0.988 & 0.989 \\
Gemma-2-27B       & 0.984 & 0.988 & 0.997 & 0.988 & 0.986 & 0.990 & 0.975 & 0.974 & 0.984 & 0.987 & 0.985 \\
Phi-4-14B         & 0.986 & 0.992 & \textbf{1.000} & 0.985 & 0.986 & 0.992 & 0.979 & 0.991 & 0.991 & 0.990 & 0.989 \\
Gemma-3-12B       & 0.989 & 0.994 & 0.997 & 0.985 & 0.984 & 0.993 & 0.992 & 0.996 & 0.985 & 0.992 & 0.989 \\
Llama-3.1-8B      & 0.976 & 0.980 & 0.997 & 0.979 & 0.978 & 0.989 & 0.979 & 0.979 & 0.976 & 0.982 & 0.979 \\
Phi-4-4B          & 0.954 & 0.969 & 0.987 & 0.959 & 0.970 & 0.980 & 0.958 & 0.979 & 0.975 & 0.973 & 0.971 \\
Gemma-3-4B        & 0.957 & 0.970 & 0.980 & 0.962 & 0.953 & 0.965 & 0.962 & 0.949 & 0.958 & 0.956 & 0.956 \\
Gemma-3-1B        & 0.651 & 0.692 & 0.647 & 0.649 & 0.619 & 0.679 & 0.732 & 0.634 & 0.658 & 0.656 & 0.664 \\
\bottomrule
\end{tabular}
\end{adjustbox}

\vspace{2pt}
\begin{minipage}{0.97\textwidth}
\scriptsize
$^{\ast}$Model used to generate the benchmark.
\end{minipage}
\end{table}

\begin{figure}[t]
    \centering
    \includegraphics[width=\linewidth]{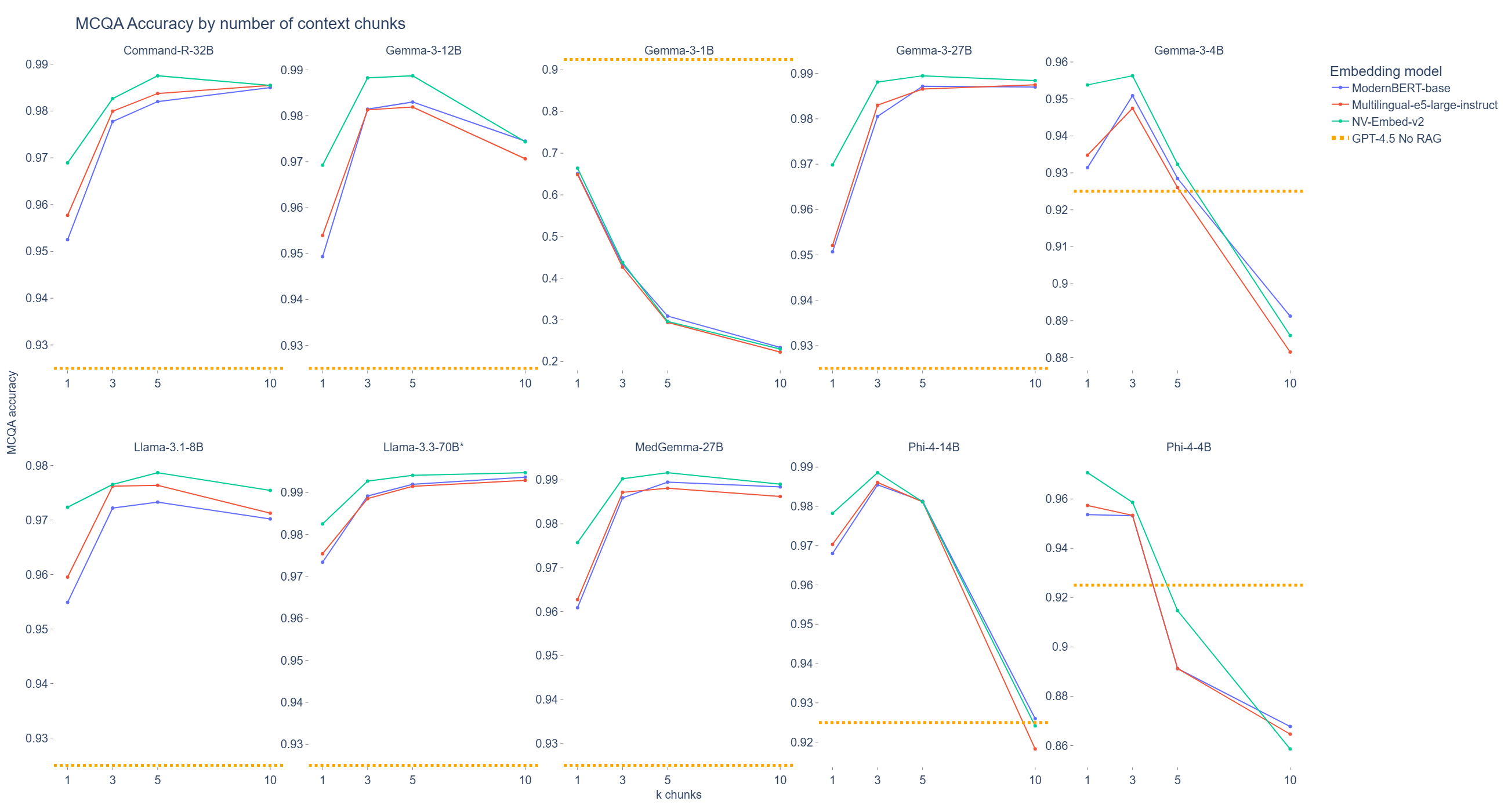}
    \caption{\textbf{MCQA accuracy by number of context chunks} The MCQA scores of generation models by the number of context chunks used in the prompt. Scores are shown for the 3 different retrieval runs used. *Model used to generate benchmark}
    \label{fig:mcqa by k}
\end{figure} 

Most models reach maximum MCQA accuracy with $k\in\{3,5\}$ context chunks, though all LLMs exceed the highest score without retrieval even at $k=1$; only \textbf{Llama-3.3-70B} continues to improve at $k=10$, and the marginal gain is small relative to the additional token cost. At $k=5$, LLMs had access to the target chunk at least 95\% of the time, and the increase in recall from $k=5$ to $k=10$ comes with additional noise from less relevant chunks. Using \textbf{NV-Embed-v2} as the retriever raises accuracy relative to the other embedding models shown. However, performance at $k=1$ is strong despite the retrieved chunk being fully relevant for only $\leq 80\%$ of queries.

\begin{figure}[t]
    \centering
    \begin{subfigure}{\textwidth}
        \label{subfig: retrieval_v_rank}
        \includegraphics[width=\textwidth]{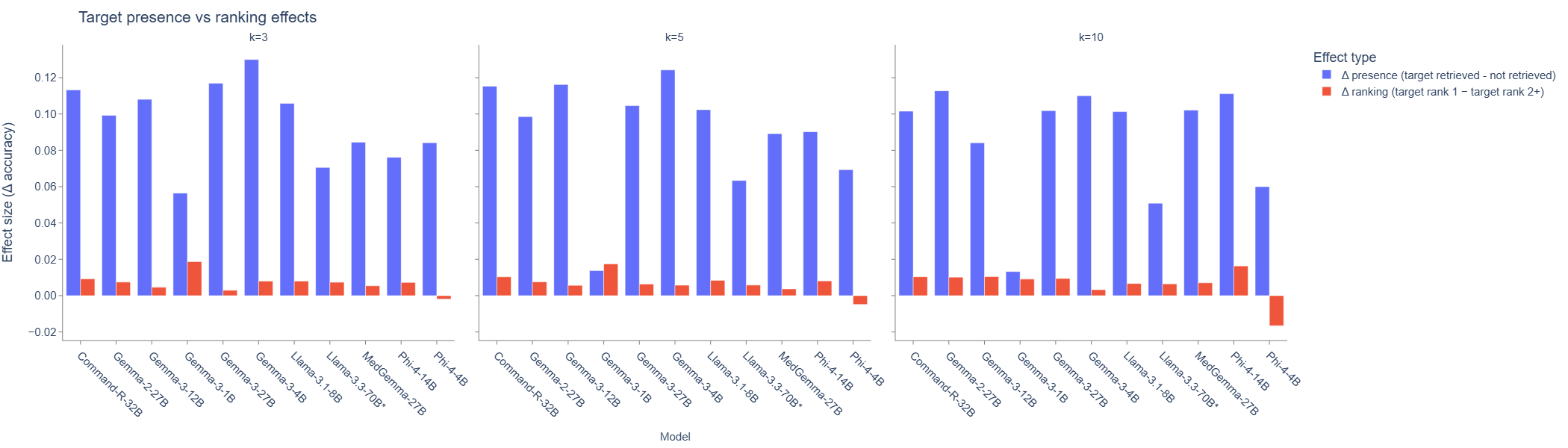}
        \subcaption{}
    \end{subfigure}
    \begin{subfigure}{\textwidth}
        {\label{subfig:rank_sensitivity}}
        \includegraphics[width=\textwidth]{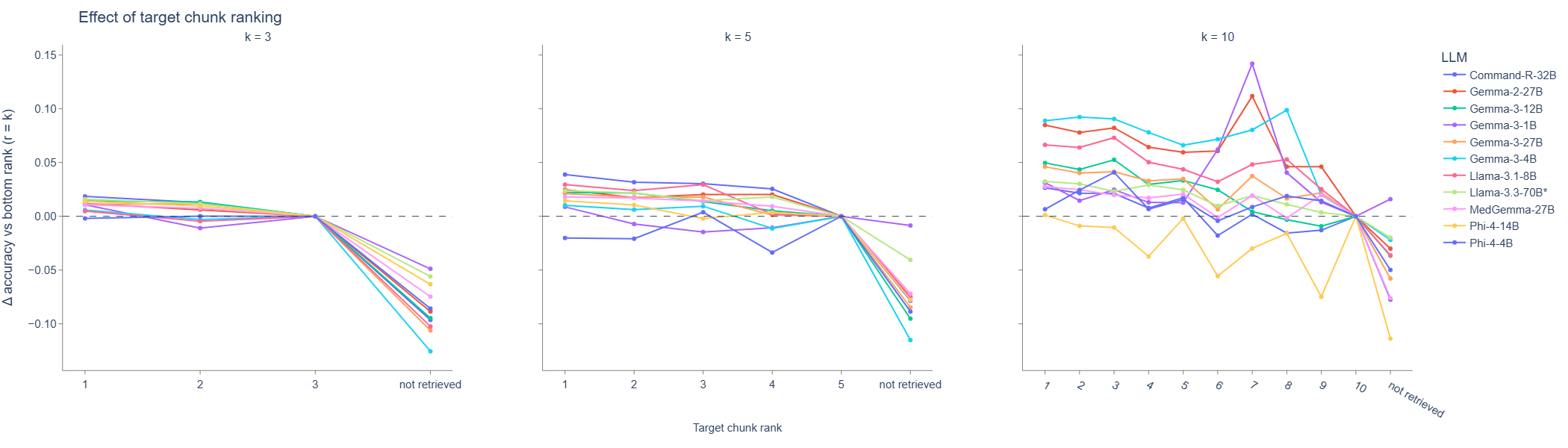}
        \subcaption{}
    \end{subfigure}
    \caption{\textbf{Effects of target chunk retrieval and rank on MCQA accuracy} (a) The differences in MCQA accuracy when target chunk retrieved vs not retrieved and retrieved at rank $r=1$ vs $r\geq2$, for $k\in\{3,5,10\}$. (b) The difference in MCQA accuracy when target chunk is retrieved at rank $r$, or not retrieved, compared to $r=k$, for $k\in\{3,5,10\}$.} 
    \label{fig:accuracy delta}
\end{figure}

The impact of retrieval quality is further shown in Figure~\ref{fig:accuracy delta}. Retrieving the correct chunk yields the largest performance gains, with ranking providing a consistent secondary benefit that is strongest when longer context sections introduce more noise. Accuracy generally declines as the target chunk moves down the ranked list, although smaller models (e.g., \textbf{Gemma-3-1B}, \textbf{Phi-4-4B}, \textbf{Gemma-3-4B}) exhibit noisier and occasionally counter-intuitive responses, including slight improvements at lower ranks, whereas larger models (e.g., \textbf{Llama-3-70B}, \textbf{MedGemma-27B}, \textbf{Gemma-3-27B}) show smoother, more consistent declines and overall reduced sensitivity to rank.

\subsubsection{LLM-as-Judge Validation}

\begin{table}[H]
\centering
\caption{Cohen's $\kappa$ and macro-F1 agreement between human reviewers and the LLM judge (GPT-OSS-120B) with 95\% bootstrap CIs.}
\label{tab:judge_agreement}
\scriptsize
\begin{adjustbox}{max width=\columnwidth}
\begin{tabular}{@{}lcccccc@{}}
\toprule
& \multicolumn{2}{c}{\textbf{Reviewer 1 vs Reviewer 2}} & \multicolumn{2}{c}{\textbf{Judge vs Reviewer 1}} & \multicolumn{2}{c}{\textbf{Judge vs Reviewer 2}} \\
\cmidrule(lr){2-3}\cmidrule(lr){4-5}\cmidrule(lr){6-7}
\textbf{Criterion} & $\boldsymbol{\kappa}$ & \textbf{macro-F1} & $\boldsymbol{\kappa}$ & \textbf{macro-F1} & $\boldsymbol{\kappa}$ & \textbf{macro-F1} \\
\midrule
Faithfulness       & 0.60 [0.43, 0.75] & 0.80 [0.71, 0.88] & 0.64 [0.46, 0.78] & 0.82 [0.73, 0.89] & 0.71 [0.54, 0.83] & 0.85 [0.77, 0.92] \\
Completeness       & 0.76 [0.58, 0.89] & 0.88 [0.79, 0.94] & 0.59 [0.40, 0.73] & 0.79 [0.69, 0.87] & 0.57 [0.38, 0.72] & 0.78 [0.68, 0.86] \\
Factual consistency& 0.46 [0.27, 0.65] & 0.73 [0.63, 0.82] & 0.06 [-0.13, 0.27] & 0.53 [0.44, 0.64] & 0.17 [-0.03, 0.37] & 0.59 [0.48, 0.69] \\
Clarity            & 0.35 [0.17, 0.51] & 0.66 [0.56, 0.75] & 0.26 [0.07, 0.44] & 0.62 [0.52, 0.72] & 0.61 [0.45, 0.76] & 0.81 [0.72, 0.88] \\
Overall            & 0.56 [0.48, 0.64] & 0.78 [0.74, 0.82] & 0.42 [0.33, 0.51] & 0.71 [0.66, 0.76] & 0.55 [0.47, 0.63] & 0.78 [0.73, 0.82] \\
\bottomrule
\end{tabular}
\end{adjustbox}
\end{table}

We validated the LLM-as-Judge rubric on a sample of 100 responses independently annotated by two human reviewers (Table~\ref{tab:judge_agreement}). The validation reveals substantial variation in how consistently different quality dimensions can be assessed, even by human raters, underscoring the difficulty of evaluating free-form responses to public health questions.

Completeness was the most reliably assessed criterion, with high inter-reviewer agreement ($\kappa \approx 0.76$; macro-F1 $\approx 0.88$) and strong judge--reviewer alignment ($\kappa \approx 0.57$--$0.59$). Faithfulness was also reasonably consistent across both human raters ($\kappa \approx 0.60$; macro-F1 $\approx 0.80$) and between the judge and reviewers ($\kappa \approx 0.6$--$0.7$). For both criteria, judge performance falls within the range of human disagreement, and at the aggregate level, judge agreement with Reviewer~2 is essentially indistinguishable from inter-reviewer agreement (overall $\kappa \approx 0.55$ vs.\ $0.56$; macro-F1 $\approx 0.78$ in both cases).

However, clarity and factual consistency proved much harder to assess reliably. Both criteria exhibited lower inter-reviewer agreement, indicating that humans themselves struggle to apply these dimensions consistently to public health responses. For factual consistency in particular, judge--human agreement was weak ($\kappa$ near zero and substantially below inter-reviewer $\kappa$), meaning that automated scoring for this criterion does not currently reproduce human judgments. Factual consistency would be a critical dimension for public health applications, yet we find it the hardest to evaluate reliably for both humans and the LLM judge. We therefore focus subsequent analysis on faithfulness and completeness, where both human agreement and judge alignment are sufficient to support robust conclusions, while recognising that developing reliable evaluation methods for factual consistency remains an important open challenge.

\subsubsection{Free-form Response Evaluations}
\begin{figure}[h]
    \centering
    \includegraphics[width=\textwidth]{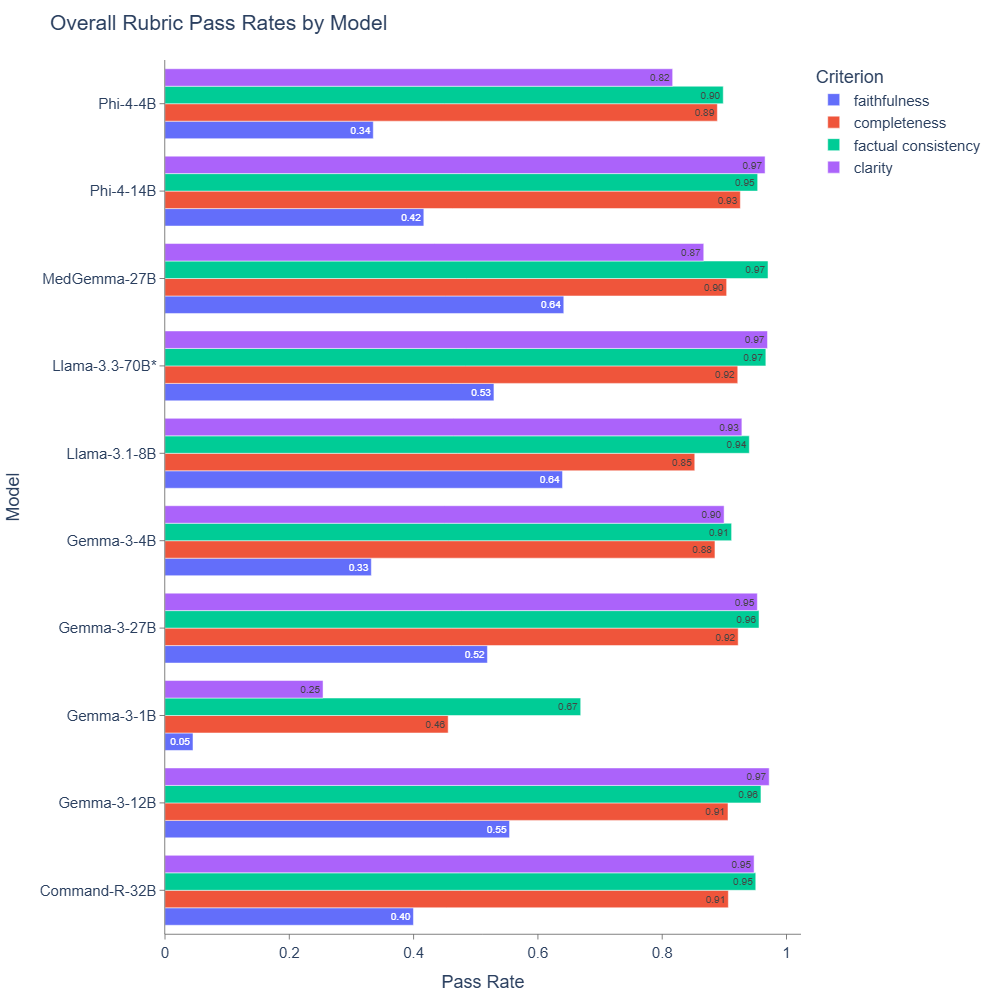}
    \caption{\textbf{Judge Rubric scores per model} Proportion of free-form answers that meet each rubric criteria for all LLMs according to the LLM Judge (GPT-OSS-120B)}
    \label{fig:rubric pass marks}
\end{figure} 

Figure \ref{fig:rubric pass marks} shows judge pass rates for free-form answers across models. Medium and large models achieve consistently high completeness and clarity (typically $\sim$0.88--0.97), indicating that they identify the key guidance needed to answer the question and present it in a legible, interpretable way even without guiding answer options. The smallest model (\textbf{Gemma-3-1B}) is a clear outlier, with substantially lower completeness and clarity. In contrast, faithfulness is the dominant failure mode across all models: even among larger models, only around half to two-thirds of responses meet the faithfulness criterion (e.g., 0.64 for the best-performing model), indicating a broad tendency to incorporate guidance from the retrieved context beyond what is strictly required to answer the question.

\begin{figure}[t]
    \centering
    \includegraphics[width=\textwidth]{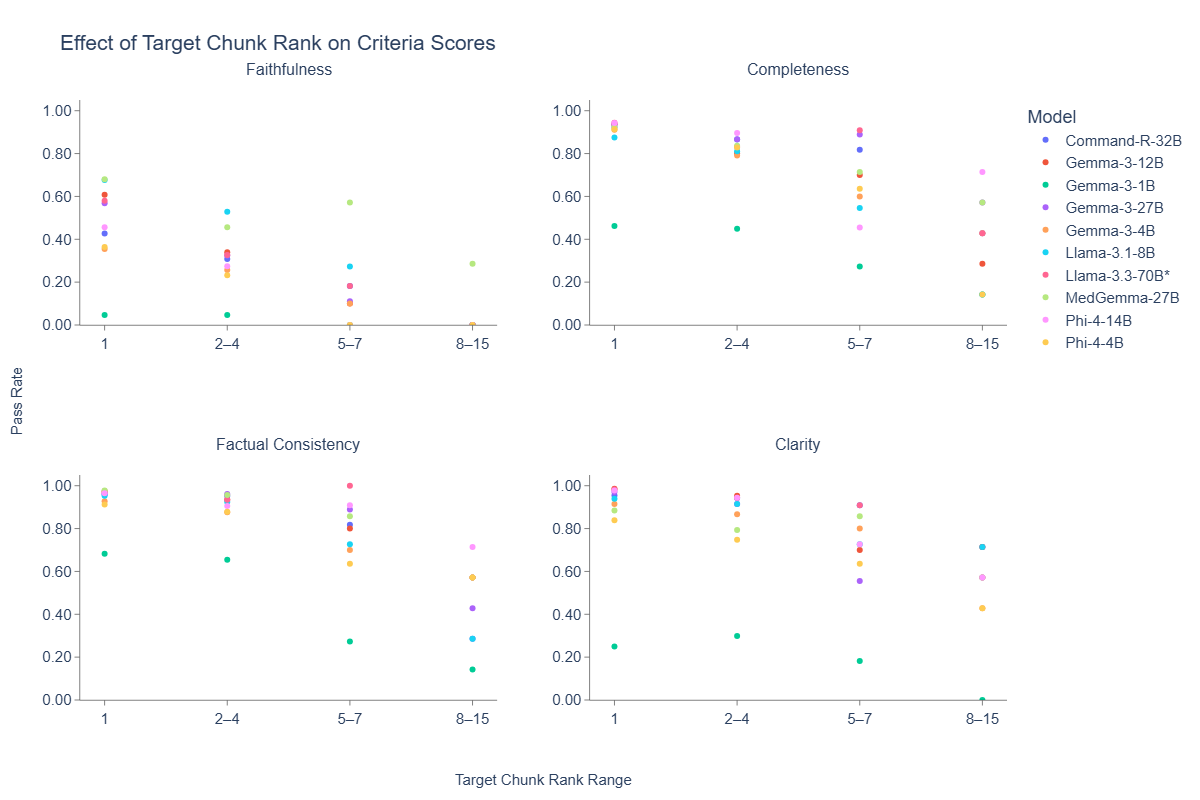}
    \caption{\textbf{Effect of target chunk rank on criteria scores} Proportion of free-form answers that meet each judge criteria when the target chunk is retrieved in rank ranges $r=1,\, 2\leq r\leq4,\,5\leq r\leq7,\, 8\leq r\leq15$.}
    \label{fig:rubric by rank}
\end{figure}

Retrieval ranking has a pronounced effect on faithfulness (Figure \ref{fig:rubric by rank}). As the target chunk rank $r$ increases, faithfulness drops sharply, consistent with models relying more heavily on earlier context and drawing in extraneous guidance when the most relevant information appears later in the prompt. By contrast, completeness and clarity show only modest degradation beyond $r \geq 5$, suggesting that once models locate the relevant guidance they can still produce a coherent answer, even if the evidence appears lower in the context.

\section{Limitations}
Our study has several limitations. First, in retrieval evaluation, relevance is treated as binary with a single target chunk per query. Many public health questions are supported by multiple passages, so single-target metrics provide only a partial view of retrieval quality and may understate the value of retrieving alternative, equally valid chunk. Second, our analysis of chunk-length effects leverages the existing chunk structure rather than systematically varying chunk size over the same underlying text, limiting causal interpretation. Third, while we observe topic-level variability in ranking quality, our experiments do not isolate the underlying drivers (e.g., topic-specific language, formatting conventions, or document structure).

Fourth, the multiple-choice format enables partial-credit behaviours such as option elimination, meaning that MCQA gains may overestimate real-world capability. This concern is reinforced by the observation that high MCQA accuracy persists even when the top-ranked chunk is not the labelled target. This gap could reflect two non-exclusive mechanisms: the retrieved chunk may contain sufficient information to eliminate incorrect options without fully answering the question, or it may contain genuinely relevant guidance that our single-target labelling scheme does not capture.

In free-form evaluation, our rubric scores are computed with respect to a single reference chunk. The judge cannot distinguish between genuinely unsupported additions and statements supported elsewhere in the corpus, so faithfulness should be interpreted as ``faithful to the provided evidence'' rather than ``faithful to the entire guidance corpus''. Additionally, weak judge--human agreement on factual consistency means that full-dataset scores for this criterion should be treated as indicative rather than definitive. Finally, because PubHealthBench questions are derived from the guidance itself, they are by construction answerable from the corpus and do not test multi-chunk reasoning or out-of-corpus queries where safe systems should acknowledge uncertainty.

\section{Discussion}
\subsection{Retrieval}
The strong performance of \textbf{NV-Embed-v2} across all metrics aligns with external evaluations where NV-Embed models top the MTEB leaderboard, reflecting deliberate optimisation for embedding and search tasks \cite{lee2024nvembed}. The universal gains from hybrid retrieval are consistent with prior work showing that combining dense and sparse signals improves recall on domain-specific tasks by pairing semantic similarity with specialised language matching. Importantly, hybrid retrieval enables small and medium-sized models to perform comparably to much larger models, making it essential when working within compute and memory constraints. Given the wide range of viable retrieval configurations in our public health guidance setting, domain-specific benchmarks such as PubHealthBench are essential for validating and selecting effective setups.

Chunk design also exerts a clear influence. The degradation of ranking quality with increasing chunk length (\S\ref{sec:ret_results}, Figure~\ref{fig:retrieval by len}) is consistent with longer chunks blurring topical boundaries, mixing multiple concepts, and reducing the discriminative power of embeddings. That LLM-generated summaries partially mitigate this effect suggests semantic compression can make relevant material easier to retrieve; however, ranking quality still drops for the longest chunks, indicating that single-shot summaries do not fully preserve the semantics of complex guidance sections. These findings suggest pairing summarisation with more principled chunking (e.g., structure-aware or topic-aware segmentation) rather than relying on summarisation alone.

The weak but systematic effect of topic on ranking quality but not recall (Figure~\ref{fig:retrival_avg_diff}) suggests that general-purpose embedding models struggle to separate the most relevant text from superficially similar content in narrow public-health subdomains. This consistency across models points to topic-associated properties of the queries and guidance chunks, such as language patterns, structure, or formatting, as the likely drivers. Where fine-grained improvements in ranking quality are needed, a further systematic exploration of how these properties individually impact retrieval is required.

\subsection{MCQA with Retrieval}
Our results demonstrate that with high-quality retrieval, small and medium-sized open-weight models reliably outperform a larger closed model used without retrieval. For knowledge-intensive QA over PubHealthBench, retrieval quality can thus compensate for reduced parameter count: it is more effective to pair a smaller LLM with a well-designed retrieval stack than to rely on a larger model operating without retrieval. The optimal context size for smaller models is $k \in \{3,5\}$ chunks; beyond that range, additional chunks introduce noise and yield diminishing or negative returns relative to the extra token cost. This constrains the useful context budget for lightweight deployments and suggests that optimisation effort should focus on what goes into those few chunks rather than simply increasing $k$.

Within this constrained context window, retrieval quality, first in terms of recall, and then ranking quality, is the primary driver of MCQA performance (Figure~\ref{fig:accuracy delta}). The interaction between rank sensitivity and model size creates a spectrum of viable design points. Smaller LLMs are more sensitive to ranking quality and work best with fewer, well-ranked chunks, so they must be paired with retrieval setups that keep relevant evidence near the top of a short list. Larger models can partially mitigate weaker ranking because they are better at extracting information from noisier contexts, making it possible to trade off retrieval quality against constraints such as latency, cost, or infrastructure limitations.

\subsection{Free-Form QA with Retrieval}
Many of our MCQA findings generalise to the more realistic free-form setting. Medium and large models generally identify relevant guidance from long retrieved contexts and express it in a clear, usable form even without multiple-choice options. The main practical challenge is controlling scope: models often draw broadly from retrieved context, producing answers that go beyond what is required and may combine multiple recommendations. In a public health setting, this matters because additional guidance, even when plausible, can reduce response precision and increase opportunities for misunderstanding. The sharp decline in faithfulness as target chunk rank increases (Figure~\ref{fig:rubric by rank}) reinforces ranking quality as a key design consideration for RAG systems in free-form public health QA.

Our human validation results inform the reliability of these automated scores. The judge shows strong alignment with humans for faithfulness and reasonable consistency for completeness, so these dimensions provide the most dependable basis for comparing retrieval setups and models at scale. By contrast, weak judge--human agreement on factual consistency and low inter-rater agreement on clarity highlight the need for further work on automated evaluation methods that more consistently reproduce human judgments of response quality in public health QA.

\subsection{Public Health Implications}
Together, these findings have several implications for the design and deployment of RAG systems in public health. In our setup, augmenting LLMs with UK public health documents is sufficient to keep the system aligned with official guidance, to a high degree of accuracy ($\sim 99\%$ MCQA accuracy), without requiring model retraining. This is consistent with recent work showing that access to up-to-date sources is a key determinant of answer quality on our public health QA benchmark, and that RAG yields more faithful and evidence-backed responses than closed-book LLMs alone \citep{Harris2025PubHealthBench, fernandezpichel2025searchenginesllmshealth, arora2025healthbenchevaluatinglargelanguage}. For public health agencies that revise guidance in response to emerging evidence or outbreaks, the ability to update the knowledge base directly is important for maintaining the accuracy of LLM applications.

Second, a well-designed retrieval stack allows small and medium-sized open-weight models to match or exceed the performance of larger models running without retrieval. This makes RAG-based systems a practical option for public health institutions operating under budget and infrastructure constraints, offering both competitive performance and greater control over factors such as cost and information update cadence.

Additionally, public health deployments bring specific data-governance requirements. Health data are highly sensitive, and regulators increasingly emphasise data minimisation, strict control of sensitive data, and careful consideration of where and how LLMs are hosted \citep{li2025implementingllmshealthcare, edpb2024llmprivacy, DHSC2025GoodPracticeDigitalHealth}. RAG is compatible with deployment patterns in which models and knowledge bases are hosted on-premise or within controlled environments, while still providing users with natural-language access to public health guidance.

\section{Conclusion}
This work extends PubHealthBench into a retrieval-augmented setting by allowing models to retrieve evidence from the same corpus of UK public health guidance used to construct the benchmark \citep{Harris2025PubHealthBench}. Using this setup, we make three key findings. First, we show that retrieval effectiveness in this public health guidance domain is highly sensitive to embedding model choice and system design: hybrid retrieval consistently improves retrieval quality, carefully crafting chunk length and uisng larger embedding models also improve relevance ranking. Second, we find that introducing retrieval of official guidance substantially improves MCQA performance, enabling smaller open-weight models to match or outperform a larger models used without retrieval, with gains driven primarily by retrieval quality and careful context selection. Third, we find that these improvements largely carry over to free-form answering, where high-quality ranking yields more focused and faithful responses, but also exposes task-design limitations that matter for real deployments (e.g., multi-evidence support and out-of-corpus queries). Further, our LLM judge evaluations do not fully match human expert assessments, indicating the need for further development in this area as robust automated evaluations are a key requirement for deploying LLM based applications in public health use cases \cite{Lekadir2025FUTUREAI}.

Overall, our results reinforce a practical conclusion for the application of RAG in public health: the retrieval layer is one of the primary determinants of QA performance and a key lever for deploying capable systems under realistic operational constraints. By grounding responses in an well maintained guidance corpus, RAG provides a pathway to systems that are easier to keep up to date as recommendations evolve, and more feasible to deploy in controlled environments where governance and data security requirements are stringent \citep{DHSC2025GoodPracticeDigitalHealth}. Future work should strengthen the benchmark’s retrieval supervision (e.g., multiple relevant chunks per query), broaden evaluation for free-form responses, and introduce harder settings that require multi-chunk reasoning and explicitly test out-of-corpus failure modes.
\begingroup
\sloppy
\bibliography{ref}

@article{Lekadir2025FUTUREAI,
  title        = {FUTURE-AI: international consensus guideline for trustworthy and deployable artificial intelligence in healthcare},
  author       = {Lekadir, Karim and Frangi, Alejandro F. and Porras, Antonio R. and Glocker, Ben and Cintas, Celia and Langlotz, Curtis P. and Weicken, Eva and Asselbergs, Folkert W. and Prior, Fred and Collins, Gary S. and Kaissis, Georgios and Tsakou, Gianna and Buvat, Ir{\`e}ne and Kalpathy-Cramer, Jayashree and Mongan, John and Schnabel, Julia A. and Kushibar, Kaisar and Riklund, Katrine and Marias, Kostas and Amugongo, Lameck M. and Fromont, Lauren A. and Maier-Hein, Lena and Cerd{\'a}-Alberich, Leonor and Mart{\'i}-Bonmat{\'i}, Luis and Cardoso, M. Jorge and Bobowicz, Maciej and Shabani, Mahsa and Tsiknakis, Manolis and Zuluaga, Maria A. and Fritzsche, Marie-Christine and Camacho, Marina and Linguraru, Marius George and Wenzel, Markus and De Bruijne, Marleen and Tolsgaard, Martin G. and Goisauf, Melanie and Cano Abad{\'i}a, M{\'o}nica and Papanikolaou, Nikolaos and Lazrak, Noussair and Pujol, Oriol and Osuala, Richard and Napel, Sandy and Colantonio, Sara and Starmans, Martijn P. A. and Joshi, Smriti and Klein, Stefan and Auss{\'o}, Susanna and Rogers, Wendy A. and Salahuddin, Zohaib and FUTURE-AI Consortium},
  journal      = {BMJ},
  year         = {2025},
  month        = feb,
  day          = {5},
  volume       = {388},
  pages        = {e081554},
  doi          = {10.1136/bmj-2024-081554},
  url          = {https://pubmed.ncbi.nlm.nih.gov/39909534/}
}

@inproceedings{nimo-etal-2025-afrimed,
    title = "{A}fri{M}ed-{QA}: A Pan-{A}frican, Multi-Specialty, Medical Question-Answering Benchmark Dataset",
    author = {Nimo, Charles  and
      Olatunji, Tobi  and
      Owodunni, Abraham Toluwase  and
      Abdullahi, Tassallah  and
      Ayodele, Emmanuel  and
      Sanni, Mardhiyah  and
      Aka, Ezinwanne C.  and
      Omofoye, Folafunmi  and
      Yuehgoh, Foutse  and
      Faniran, Timothy  and
      Dossou, Bonaventure F. P.  and
      Yekini, Moshood O.  and
      Kemp, Jonas  and
      Heller, Katherine A  and
      Omeke, Jude Chidubem  and
      Md, Chidi Asuzu  and
      Etori, Naome A  and
      Ndiaye, A{\"i}m{\'e}rou  and
      Okoh, Ifeoma  and
      Ocansey, Evans Doe  and
      Kinara, Wendy  and
      Best, Michael L.  and
      Essa, Irfan  and
      Moore, Stephen Edward  and
      Fourie, Chris  and
      Asiedu, Mercy Nyamewaa},
    editor = "Che, Wanxiang  and
      Nabende, Joyce  and
      Shutova, Ekaterina  and
      Pilehvar, Mohammad Taher",
    booktitle = "Proceedings of the 63rd Annual Meeting of the Association for Computational Linguistics (Volume 1: Long Papers)",
    month = jul,
    year = "2025",
    address = "Vienna, Austria",
    publisher = "Association for Computational Linguistics",
    url = "https://aclanthology.org/2025.acl-long.96/",
    doi = "10.18653/v1/2025.acl-long.96",
    pages = "1948--1973",
    isbn = "979-8-89176-251-0"
}

@article{Harris2025PubHealthBench,
  title   = {Healthy LLMs? Benchmarking LLM Knowledge of UK Government Public Health Information},
  author  = {Harris, Joshua and Grayson, Fan and Feldman, Felix and Laurence, Timothy and Nonnenmacher, Toby and Higgins, Oliver and Loman, Leo and Patel, Selina and Finnie, Thomas and Collins, Samuel and Borowitz, Michael},
  journal = {arXiv preprint arXiv:2505.06046},
  year    = {2025},
  url     = {https://arxiv.org/abs/2505.06046}
}

@article{Ventura2025HealthQA-BR,
  title   = {HealthQA-BR: A System-Wide Benchmark Reveals Critical Knowledge Gaps in Large Language Models},
  author  = {Ventura D'Addario, Andrew Maranh{\~a}o},
  journal = {arXiv preprint arXiv:2506.21578},
  year    = {2025},
  url     = {https://arxiv.org/abs/2506.21578}
}

@article{Yan2024MedLLM-Benchmarks,
  title   = {Large Language Model Benchmarks in Medical Tasks},
  author  = {Yan, Lawrence K. Q. and Niu, Qian and Li, Ming and Zhang, Yichao and Yin, Caitlyn H. and Fei, Cheng and Peng, Benji and Bi, Ziqian and Feng, Pohsun and others},
  journal = {arXiv preprint arXiv:2410.21348},
  year    = {2024},
  url     = {https://arxiv.org/abs/2410.21348}
}

@article{Shi2024MKRAG,
  title   = {MKRAG: Medical Knowledge Retrieval Augmented Generation for Medical Question Answering},
  author  = {Shi, Yucheng and Xu, Shaochen and Yang, Tianze and Liu, Zhengliang and Liu, Tianming and Li, Quanzheng and Li, Xiang and Liu, Ninghao},
  journal = {arXiv preprint arXiv:2309.16035},
  year    = {2024},
  url     = {https://arxiv.org/abs/2309.16035}
}

@article{Borgeaud2022RETRO,
  title   = {Improving language models by retrieving from trillions of tokens},
  author  = {Borgeaud, Sebastian and Mensch, Arthur and Hoffmann, Jordan and Cai, Trevor and Rutherford, Eliza and Millican, Katie and van den Driessche, George and Lespiau, Jean-Baptiste and Damoc, Bogdan and Clark, Aidan and de Las Casas, Diego and Guy, Aurelia and Menick, Jacob and Ring, Roman and Hennigan, Tom and Huang, Saffron and Maggiore, Loren and Jones, Chris and Cassirer, Albin and Brock, Andy and Paganini, Michela and Irving, Geoffrey and Vinyals, Oriol and Osindero, Simon and Simonyan, Karen and Rae, Jack W. and Elsen, Erich and Sifre, Laurent},
  journal = {arXiv preprint arXiv:2112.04426},
  year    = {2022},
  url     = {https://arxiv.org/abs/2112.04426}
}

@article{Shao2024DatastoreScaling,
  title   = {Scaling Retrieval-Based Language Models with a Trillion-Token Datastore},
  author  = {Shao, Rulin and He, Jacqueline and Asai, Akari and Shi, Weijia and Dettmers, Tim and Min, Sewon and Zettlemoyer, Luke and Koh, Pang Wei},
  journal = {arXiv preprint arXiv:2407.12854},
  year    = {2024},
  url     = {https://arxiv.org/abs/2407.12854}
}

@article{Boeckling2024WalkRetrieve,
  title   = {Walk\&Retrieve: Simple Yet Effective Zero-shot Retrieval-Augmented Generation via Knowledge Graph Walks},
  author  = {B{\"o}ckling, Martin and Paulheim, Heiko and Iana, Andreea},
  journal = {arXiv preprint arXiv:2505.16849},
  year    = {2025},
  url     = {https://arxiv.org/abs/2505.16849}
}

@article{hosseini2024medicalQA,
  title   = {A Benchmark for Long-Form Medical Question Answering},
  author  = {Hosseini, Pedram and Sin, Jessica M. and Ren, Bing and Thomas, Bryceton G. and Nouri, Elnaz and Farahanchi, Ali and Hassanpour, Saeed},
  journal = {arXiv preprint arXiv:2411.09834},
  year    = {2024},
  url     = {https://arxiv.org/abs/2411.09834}
}

@inproceedings{manes2023kqa,
  title     = {K-QA: A Real-World Medical Q\&A Benchmark},
  author    = {Manes, Itay and Ronn, Naama and Cohen, David and Ilan Ber, Ran and Horowitz-Kugler, Zehavi and Stanovsky, Gabriel},
  booktitle = {Proceedings of the 23rd Workshop on Biomedical NLP (BioNLP 2023)},
  pages     = {277--294},
  year      = {2023}
}

@article{singh2025mcq,
  title   = {It is Too Many Options: Pitfalls of Multiple-Choice Questions in Generative AI and Medical Education},
  author  = {Singh, Shrutika and Alyakin, Anton and Alber, Daniel A. and Stryker, Jaden and Hernandez-Rovira, Miguel and Park, Ki Yun and Oermann, Eric K. and others},
  journal = {arXiv preprint arXiv:2503.13508},
  year    = {2025},
  url     = {https://arxiv.org/abs/2503.13508}
}

@inproceedings{xu2023lfqa-eval,
  title     = {A Critical Evaluation of Evaluations for Long-form Question Answering},
  author    = {Xu, Fangyuan and Song, Yixiao and Iyyer, Mohit and Choi, Eunsol},
  booktitle = {Proceedings of the 61st Annual Meeting of the Association for Computational Linguistics (ACL)},
  pages     = {3225--3245},
  year      = {2023}
}

@inproceedings{xian2025lfqa,
  title     = {An Empirical Study of Evaluating Long-form Question Answering},
  author    = {Xian, Ning and Fan, Yixing and Zhang, Ruqing and de Rijke, Maarten and Guo, Jiafeng},
  booktitle = {Proceedings of the 48th International ACM SIGIR Conference on Research and Development in Information Retrieval},
  year      = {2025},
  doi       = {10.1145/3726302.3729895}
}

@article{Zheng2023MTBench,
  title   = {Judging LLM-as-a-Judge with MT-Bench and Chatbot Arena},
  author  = {Zheng, Lianmin and Chiang, Wei-Lin and Sheng, Ying and Zhuang, Siyuan and Wu, Zhanghao and Zhuang, Yonghao and Lin, Zi and Li, Zhuohan and Li, Dacheng and Xing, Eric P. and Zhang, Hao and Gonzalez, Joseph E. and Stoica, Ion},
  journal = {arXiv preprint arXiv:2306.05685},
  year    = {2023},
  url     = {https://arxiv.org/abs/2306.05685}
}

@misc{fang2024scalinglawsdenseretrieval,
  title         = {Scaling Laws For Dense Retrieval},
  author        = {Fang, Yan and Zhan, Jingtao and Ai, Qingyao and Mao, Jiaxin and Su, Weihang and Chen, Jia and Liu, Yiqun},
  year          = {2024},
  eprint        = {2403.18684},
  archivePrefix = {arXiv},
  primaryClass  = {cs.IR},
  url           = {https://arxiv.org/abs/2403.18684}
}

@misc{GTE-ModernColBERT,
  title  = {GTE-ModernColBERT},
  author = {Chaffin, Antoine},
  year   = {2025},
  url    = {https://huggingface.co/lightonai/GTE-ModernColBERT-v1}
}

@misc{wu2024medicalgraphragsafe,
  title         = {Medical Graph RAG: Towards Safe Medical Large Language Model via Graph Retrieval-Augmented Generation},
  author        = {Wu, Junde and Zhu, Jiayuan and Qi, Yunli and Chen, Jingkun and Xu, Min and Menolascina, Filippo and Grau, Vicente},
  year          = {2024},
  eprint        = {2408.04187},
  archivePrefix = {arXiv},
  primaryClass  = {cs.CV},
  url           = {https://arxiv.org/abs/2408.04187}
}

@article{Sharma2025RAGSurvey,
  title        = {Retrieval-Augmented Generation: A Comprehensive Survey of Architectures, Enhancements, and Robustness Frontiers},
  author       = {Sharma, Chaitanya},
  journal      = {arXiv preprint arXiv:2506.00054},
  year         = {2025},
  url          = {https://arxiv.org/abs/2506.00054}
}

@article{Vach2025NeuroRAG,
  title   = {Evaluating Retrieval-Augmented Generation–enhanced Large Language Models for Question Answering On German Neurovascular Guidelines},
  author  = {Vach, Marius and Gliem, Michael and Weiss, Daniel and Ivan, Vivien Lorena and Hauke, Frederik and Boschenriedter, Christian and Rubbert, Christian and Caspers, Julian},
  journal = {Clinical Neuroradiology},
  year    = {2025},
  doi     = {10.1007/s00062-025-01562-z},
  note    = {(Online ahead of print)}
}

@article{Pratapa2025EstimatingContext,
  title        = {Estimating Optimal Context Length for Hybrid Retrieval-Augmented Multi-document Summarization},
  author       = {Pratapa, Adithya and Mitamura, Teruko},
  journal      = {arXiv preprint arXiv:2504.12972},
  year         = {2025},
  url          = {https://arxiv.org/abs/2504.12972}
}

@inproceedings{jin2025longcontext,
  title     = {Long-Context LLMs Meet RAG: Overcoming Challenges for Long Inputs in RAG},
  author    = {Jin, Bowen and Yoon, Jinsung and Han, Jiawei and Arik, Sercan O.},
  booktitle = {The Thirteenth International Conference on Learning Representations (ICLR)},
  year      = {2025},
  url       = {https://openreview.net/forum?id=oU3tpaR8fm}
}

@article{Kim2023Prometheus,
  title   = {Prometheus: Inducing Fine-grained Evaluation Capability in Language Models},
  author  = {Kim, Seungone and Shin, Jamin and Cho, Yejin and Jang, Joel and Longpre, Shayne and Lee, Hwaran and Yun, Sangdoo and Shin, Seongjin and Kim, Sungdong and Thorne, James and Seo, Minjoon},
  journal = {arXiv preprint arXiv:2310.08491},
  year    = {2023},
  doi     = {10.48550/arXiv.2310.08491},
  url     = {https://arxiv.org/abs/2310.08491}
}

@article{Kim2024Prometheus2,
  title   = {Prometheus 2: An Open Source Language Model Specialized in Evaluating Other Language Models},
  author  = {Kim, Seungone and Shin, Jamin and Cho, Yejin and Jang, Joel and Longpre, Shayne and Lee, Hwaran and Yun, Sangdoo and Shin, Seongjin and Kim, Sungdong and Thorne, James and Seo, Minjoon},
  journal = {arXiv preprint arXiv:2405.01535},
  year    = {2024},
  url     = {https://arxiv.org/abs/2405.01535}
}

@article{Gu2024LLMJudgeSurvey,
  title   = {A Survey on LLM-as-a-Judge},
  author  = {Gu, Jiahui and others},
  journal = {arXiv preprint arXiv:2411.15594},
  year    = {2024},
  url     = {https://arxiv.org/abs/2411.15594}
}

@article{Croxford2025MedLLMJudge,
  title   = {Automating Evaluation of AI Text Generation in Healthcare using Medical LLM-as-a-Judge},
  author  = {Croxford, Euan and others},
  journal = {medRxiv},
  year    = {2025},
  url     = {https://www.medrxiv.org/content/10.1101/2025.04.22.25326219v2},
  note    = {Version v2; PubMed PMID: 40313300}
}

@misc{dangi2025evaluationgpt4ogpt4ominisvision,
  title         = {Evaluation of GPT-4o and GPT-4o-mini’s Vision Capabilities for Compositional Analysis from Dried Solution Drops},
  author        = {Dangi, Deven B. and Dangi, Beni B. and Steinbock, Oliver},
  year          = {2025},
  eprint        = {2412.10587},
  archivePrefix = {arXiv},
  primaryClass  = {cs.CV},
  url           = {https://arxiv.org/abs/2412.10587}
}

@misc{google2025embeddinggemma_blog,
  author       = {Choi, Min and Dua, Sahil and Lisak, Alice and Google DeepMind},
  title        = {Introducing EmbeddingGemma: The Best-in-Class Open Model for On-Device Embeddings},
  year         = {2025},
  month        = {September 4},
  howpublished = {Google Developer Blog},
  url          = {https://developers.googleblog.com/en/introducing-embeddinggemma/}
}

@article{lee2024nvembed,
  title   = {NV-Embed: Improved Techniques for Training LLMs as Generalist Embedding Models},
  author  = {Lee, Chankyu and Roy, Rajarshi and Xu, Mengyao and Raiman, Jonathan and Shoeybi, Mohammad and Catanzaro, Bryan and Ping, Wei},
  journal = {arXiv preprint arXiv:2405.17428},
  year    = {2024},
  note    = {Includes NV-Embed-v1 and NV-Embed-v2},
  url     = {https://arxiv.org/abs/2405.17428}
}

@article{wang2024multilingualE5,
  title   = {Multilingual E5 Text Embeddings: A Technical Report},
  author  = {Wang, Liang and Yang, Nan and Huang, Xiaolong and Yang, Linjun and Majumder, Rangan and Wei, Furu},
  journal = {arXiv preprint arXiv:2402.05672},
  year    = {2024},
  note    = {Includes the {\tt multilingual-E5-large} model},
  url     = {https://arxiv.org/abs/2402.05672}
}

@misc{meng2024sfr_embedding_mistral,
  title        = {SFR-Embedding-Mistral: Enhance Text Retrieval with Transfer Learning},
  author       = {Meng, Rui and Liu, Ye and Joty, Shafiq Rayhan and Xiong, Caiming and Zhou, Yingbo and Yavuz, Semih},
  howpublished = {Salesforce AI Research Blog / Hugging Face model card},
  year         = {2024},
  month        = {October},
  url           = {https://huggingface.co/Salesforce/SFR-Embedding-Mistral}
}

@misc{modernbert2024,
  title         = {Smarter, Better, Faster, Longer: A Modern Bidirectional Encoder for Fast, Memory Efficient, and Long Context Finetuning and Inference},
  author        = {Warner, Benjamin and Chaffin, Antoine and Clavi{\'e}, Benjamin and Weller, Orion and Hallstr{\"o}m, Oskar and Taghadouini, Said and Gallagher, Alexis and Biswas, Raja and Ladhak, Faisal and Aarsen, Tom and Cooper, Nathan and Adams, Griffin and Howard, Jeremy and Poli, Iacopo},
  year          = {2024},
  eprint        = {2412.13663},
  archivePrefix = {arXiv},
  primaryClass  = {cs.CL},
  url           = {https://arxiv.org/abs/2412.13663}
}

@misc{openai2024textembedding3large,
  title        = {text-embedding-3-large: OpenAI’s Next-Generation Large Embedding Model},
  author       = {OpenAI},
  year         = {2024},
  month        = {January},
  howpublished = {OpenAI API / Documentation},
  note         = {Creates embeddings with up to 3072 dimensions},
  url          = {https://platform.openai.com/docs/models/text-embedding-3-large}
}

@article{grattafiori2024llama3herd,
  title   = {The Llama 3 Herd of Models},
  author  = {Grattafiori, A. and others},
  journal = {arXiv preprint arXiv:2407.21783},
  year    = {2024},
  note    = {Describes the Llama 3 family, including the 70B instruction-tuned model},
  url     = {https://arxiv.org/abs/2407.21783}
}

@article{abdin2024phi4,
  title   = {Phi-4 Technical Report},
  author  = {Abdin, Marah and Aneja, Jyoti and Behl, Harkirat and Bubeck, S{\'e}bastien and Eldan, Ronen and Gunasekar, Suriya and Harrison, Michael and Hewett, Russell J. and Javaheripi, Mojan and Kauffmann, Piero and Lee, James R. and Lee, Yin Tat and Li, Yuanzhi and Liu, Weishung and Mendes, Caio C. T. and Nguyen, Anh and Price, Eric and de Rosa, Gustavo and Saarikivi, Olli and Salim, Adil and Shah, Shital and Wang, Xin and Ward, Rachel and Wu, Yue and Yu, Dingli and Zhang, Cyril and Zhang, Yi},
  journal = {arXiv preprint arXiv:2412.08905},
  year    = {2024},
  note    = {14B-parameter language model},
  url     = {https://arxiv.org/abs/2412.08905}
}

@article{gemma32025,
  title   = {Gemma 3: A Multimodal, Multilingual, Long-Context Open Model Family},
  author  = {{Gemma Team, Google DeepMind}},
  journal = {arXiv preprint arXiv:2503.19786},
  year    = {2025},
  url     = {https://arxiv.org/abs/2503.19786}
}

@article{sellergren2025medgemma,
  title   = {MedGemma: A Multimodal Generative Model Family for Medical Text and Image Comprehension},
  author  = {Sellergren, Andrew and Kazemzadeh, Sahar and Jaroensri, Tiam and Kiraly, Atilla and Traverse, Madeleine and Kohlberger, Timo and Xu, Shawn and Jamil, Fayaz and Hughes, C{\'i}an and Lau, Charles and others},
  journal = {arXiv preprint arXiv:2507.05201},
  year    = {2025},
  url     = {https://arxiv.org/abs/2507.05201}
}

@misc{cohere2024command_r_v01,
  title        = {Cohere Labs Command-R (c4ai-command-r-v01)},
  author       = {Cohere Labs / CohereForAI},
  year         = {2024},
  month        = {March},
  howpublished = {Hugging Face model card},
  note         = {35B parameter generative model; 128K token context},
  url          = {https://huggingface.co/CohereLabs/c4ai-command-r-v01}
}

@article{thakur2021beir,
  title   = {BEIR: A Heterogenous Benchmark for Zero-shot Evaluation of Information Retrieval Models},
  author  = {Thakur, Nandan and Reimers, Nils and R{\"u}ckl{\'e}, Andreas and Srivastava, Abhishek and Gurevych, Iryna},
  journal = {arXiv preprint arXiv:2104.08663},
  year    = {2021},
  doi     = {10.48550/arXiv.2104.08663},
  url     = {https://arxiv.org/abs/2104.08663}
}

@inproceedings{jeunen2024ndcg,
  author    = {Jeunen, Olivier and Potapov, Ivan and Ustimenko, Aleksei},
  title     = {On (Normalised) Discounted Cumulative Gain as an Off-Policy Evaluation Metric for Top-$n$ Recommendation},
  booktitle = {Proceedings of the 30th ACM SIGKDD Conference on Knowledge Discovery and Data Mining (KDD '24)},
  year      = {2024},
  publisher = {Association for Computing Machinery},
  doi       = {10.1145/3637528.3671687},
  url       = {https://dl.acm.org/doi/10.1145/3637528.3671687}
}

@article{wang2024birco,
  title   = {BIRCO: A Benchmark of Information Retrieval Tasks with Complex Objectives},
  author  = {Wang, Xiaoyue and Wang, Jianyou and Cao, Weili and Wang, Kaicheng and Paturi, Ramamohan and Bergen, Leon},
  journal = {arXiv preprint arXiv:2402.14151},
  year    = {2024},
  doi     = {10.48550/arXiv.2402.14151},
  url     = {https://arxiv.org/abs/2402.14151}
}

@misc{singhal2022largelanguagemodelsencode,
  title         = {Large Language Models Encode Clinical Knowledge},
  author        = {Singhal, Karan and Azizi, Shekoofeh and Tu, Tao and Mahdavi, S. Sara and Wei, Jason and Chung, Hyung Won and Scales, Nathan and Tanwani, Ajay and Cole-Lewis, Heather and Pfohl, Stephen and Payne, Perry and Seneviratne, Martin and Gamble, Paul and Kelly, Chris and Scharli, Nathaneal and Chowdhery, Aakanksha and Mansfield, Philip and Aguera y Arcas, Blaise and Webster, Dale and Corrado, Greg S. and Matias, Yossi and Chou, Katherine and Gottweis, Juraj and Tomasev, Nenad and Liu, Yun and Rajkomar, Alvin and Barral, Joelle and Semturs, Christopher and Karthikesalingam, Alan and Natarajan, Vivek},
  year          = {2022},
  eprint        = {2212.13138},
  archivePrefix = {arXiv},
  primaryClass  = {cs.CL},
  url           = {https://arxiv.org/abs/2212.13138}
}

@misc{arora2025healthbenchevaluatinglargelanguage,
  title         = {HealthBench: Evaluating Large Language Models Towards Improved Human Health},
  author        = {Arora, Rahul K. and Wei, Jason and Soskin Hicks, Rebecca and Bowman, Preston and Quiñonero-Candela, Joaquin and Tsimpourlas, Foivos and Sharman, Michael and Shah, Meghan and Vallone, Andrea and Beutel, Alex and Heidecke, Johannes and Singhal, Karan},
  year          = {2025},
  eprint        = {2505.08775},
  archivePrefix = {arXiv},
  primaryClass  = {cs.CL},
  url           = {https://arxiv.org/abs/2505.08775}
}

@InProceedings{pmlr-v174-pal22a,
  title =        {MedMCQA: A Large-scale Multi-Subject Multi-Choice Dataset for Medical domain Question Answering},
  author =       {Pal, Ankit and Umapathi, Logesh Kumar and Sankarasubbu, Malaikannan},
  booktitle =    {Proceedings of the Conference on Health, Inference, and Learning},
  pages =         {248--260},
  year =          {2022},
  editor =        {Flores, Gerardo and Chen, George H and Pollard, Tom and Ho, Joyce C and Naumann, Tristan},
  volume =        {174},
  series =        {Proceedings of Machine Learning Research},
  month =         {07--08 Apr},
  publisher =   {PMLR},
  url =          {https://proceedings.mlr.press/v174/pal22a.html}
}

@article{Jin2021WhatDD,
  title   = {What disease does this patient have? A large-scale open domain question answering dataset from medical exams},
  author  = {Di Jin and Pan, Eileen and Oufattole, Nassim and Weng, Weihuong and Fang, Hanyi and Szolovits, Peter},
  journal = {Applied Sciences},
  volume  = {11},
  number  = {14},
  pages   = {6421},
  year    = {2021}
}

@article{Ni2025TrustworthyRAGSurvey,
  title        = {Towards Trustworthy Retrieval Augmented Generation for Large Language Models: A Survey},
  author       = {Bo Ni and Zheyuan Liu and Leyao Wang and Yongjia Lei and Yuying Zhao and Xueqi Cheng and Qingkai Zeng and Luna Dong and Yinglong Xia and Krishnaram Kenthapadi and Ryan Rossi and Franck Dernoncourt and Md Mehrab Tanjim and Nesreen Ahmed and Xiaorui Liu and Wenqi Fan and Erik Blasch and Yu Wang and Meng Jiang and Tyler Derr},
  journal      = {arXiv preprint arXiv:2502.06872},
  year         = {2025},
  url          = {https://arxiv.org/abs/2502.06872}
}

@inproceedings{Li2025EnhancingRAGBestPractices,
  title        = {Enhancing Retrieval-Augmented Generation: A Study of Best Practices},
  author       = {Siran Li and Linus Stenzel and Carsten Eickhoff and Seyed Ali Bahrainian},
  booktitle    = {Proceedings of the 31st International Conference on Computational Linguistics (COLING ’25)},
  year         = {2025},
  url          = {https://aclanthology.org/2025.coling-main.449/}
}

@article{Qin2025AdaptiveMemoryRAG,
  title        = {Towards Adaptive Memory-Based Optimization for Enhanced Retrieval-Augmented Generation},
  author       = {Qitao Qin and Yucong Luo and Yihang Lu and Zhibo Chu and Xianwei Meng},
  journal      = {arXiv preprint arXiv:2504.05312},
  year         = {2025},
  url          = {https://arxiv.org/abs/2504.05312}
}

@article{FernandezPichel2025SearchEnginesLLMsHealth,
  title        = {Evaluating Search Engines and Large Language Models for Answering Health Questions},
  author       = {Marcos Fernández-Pichel and Juan C. Pichel and David E. Losada},
  journal      = {npj Digital Medicine},
  volume       = {8},
  number       = {1},
  pages        = {153},
  year         = {2025},
  doi          = {10.1038/s41746-025-01546-W},
  url          = {https://doi.org/10.1038/s41746-025-01546-W}
}

@article{Cheng2025KnowledgeOrientedRAGSurvey,
  title        = {A Survey on Knowledge-Oriented Retrieval-Augmented Generation},
  author       = {Mingyue Cheng and Yucong Luo and Jie Ouyang and Qi Liu and Huijie Liu and Li Li and Shuo Yu and Bohou Zhang and Jiawei Cao and Jie Ma and Daoyu Wang},
  journal      = {arXiv preprint arXiv:2503.10677},
  year         = {2025},
  url          = {https://arxiv.org/abs/2503.10677}
}

@inproceedings{Cormack2009ReciprocalRankFusion,
  author    = {Gordon V. Cormack and Charles L. A. Clarke and Stefan B{\"u}ttcher},
  title     = {Reciprocal Rank Fusion outperforms Condorcet and individual Rank Learning Methods},
  booktitle = {Proceedings of the 32nd Annual International ACM SIGIR Conference on Research and Development in Information Retrieval (SIGIR ’09)},
  pages     = {758--759},
  year      = {2009},
  publisher = {ACM},
  doi       = {10.1145/1571941.1572114},
  url       = {https://doi.org/10.1145/1571941.1572114}
}

@article{Bruch2023AnalysisFusionHybridRetrieval,
  author    = {Sebastian Bruch and Siyu Gai and Amir Ingber},
  title     = {An Analysis of Fusion Functions for Hybrid Retrieval},
  journal   = {ACM Transactions on Information Systems},
  volume    = {42},
  number    = {1},
  pages     = {1--35},
  year      = {2023},
  doi       = {10.1145/3596512},
  url       = {https://doi.org/10.1145/3596512}
}

@article{lewis2020rag,
  title={Retrieval-Augmented Generation for Knowledge-Intensive NLP Tasks},
  author={Lewis, Patrick and Perez, Ethan and Piktus, Aleksandra and Petroni, Fabio and Karpukhin, Vladimir and Goyal, Naman and K{\"u}ttler, Heinrich and Lewis, Mike and Yih, Wen-tau and Rockt{\"a}schel, Tim and Riedel, Sebastian and Kiela, Douwe},
  journal={Advances in Neural Information Processing Systems (NeurIPS)},
  year={2020},
  url={https://proceedings.neurips.cc/paper/2020/file/6b493230205f780e1bc26945df7481e5-Paper.pdf}
}

@article{gao2024ragsurvey,
  title={Retrieval-Augmented Generation for Large Language Models: A Survey},
  author={Gao, Yunfan and Xiong, Yun and Gao, Xinyu and Jia, Kangxiang and Pan, Jinliu and Bi, Yuxi and Dai, Yi and Sun, Jiawei and others},
  journal={arXiv preprint arXiv:2312.10997},
  year={2023},
  url={https://arxiv.org/abs/2312.10997}
}

@article{li2025implementingllmshealthcare,
  title   = {Implementing Large Language Models in Health Care: Clinician-Focused Review With Interactive Guideline},
  author  = {Li, HongYi and Fu, Jun-Fen and Python, Andre},
  journal = {Journal of Medical Internet Research},
  year    = {2025},
  volume  = {27},
  number  = {1},
  pages   = {e71916},
  doi     = {10.2196/71916},
  url     = {https://www.jmir.org/2025/1/e71916}
}

@techreport{edpb2024llmprivacy,
  title        = {Opinion 28/2024 on Certain Data Protection Aspects Related to the Processing of Personal Data in the Context of AI Models},
  author       = {{European Data Protection Board}},
  institution  = {European Data Protection Board},
  year         = {2024},
  number       = {28/2024},
  type         = {EDPB Opinion},
  month        = dec,
  address      = {Brussels, Belgium},
  url          = {https://www.edpb.europa.eu/system/files/2024-12/edpb_opinion_202428_ai-models_en.pdf},
  note         = {Adopted on 17 December 2024}
}

@misc{DHSC2025GoodPracticeDigitalHealth,
  title        = {A guide to good practice for digital and data-driven health technologies},
  author       = {{Department of Health and Social Care} and {Government Digital Service}},
  year         = {2025},
  howpublished = {\url{https://www.gov.uk/data-ethics-guidance/a-guide-to-good-practice-for-digital-and-data-driven-health-technologies}},
  note         = {Published 27 January 2025; accessed 9 December 2025}
}

@misc{Editor_2025, title={How we are pioneering artificial intelligence applications in public health }, url={https://ukhsa.blog.gov.uk/2025/03/14/how-we-are-pioneering-artificial-intelligence-applications-in-public-health/}, journal={UK Health Security Agency}, publisher={UKHSA}, author={Editor, Blog}, year={2025}, month={Mar}}

@misc{harris2025evaluatinglargelanguagemodels,
      title={Evaluating Large Language Models for Public Health Classification and Extraction Tasks}, 
      author={Joshua Harris and Timothy Laurence and Leo Loman and Fan Grayson and Toby Nonnenmacher and Harry Long and Loes WalsGriffith and Amy Douglas and Holly Fountain and Stelios Georgiou and Jo Hardstaff and Kathryn Hopkins and Y-Ling Chi and Galena Kuyumdzhieva and Lesley Larkin and Samuel Collins and Hamish Mohammed and Thomas Finnie and Luke Hounsome and Michael Borowitz and Steven Riley},
      year={2025},
      eprint={2405.14766},
      archivePrefix={arXiv},
      primaryClass={cs.CL},
      url={https://arxiv.org/abs/2405.14766}, 
}

@article{Sutton2020CDSS,
  title        = {An overview of clinical decision support systems: benefits, risks, and strategies for success},
  author       = {Sutton, Reed T. and Pincock, David and Baumgart, Daniel C. and Sadowski, Daniel C. and Fedorak, Richard N. and Kroeker, Karen I.},
  journal      = {NPJ Digital Medicine},
  year         = {2020},
  month        = feb,
  day          = {6},
  volume       = {3},
  pages        = {17},
  doi          = {10.1038/s41746-020-0221-y},
  url          = {https://pubmed.ncbi.nlm.nih.gov/32047862/}
}

@techreport{ECDC2011EvidenceBasedPublicHealth,
  title        = {Evidence-based methodologies for public health: How to assess the best available evidence when time is limited and there is lack of sound evidence},
  author       = {{European Centre for Disease Prevention and Control}},
  institution  = {European Centre for Disease Prevention and Control (ECDC)},
  year         = {2011},
  month        = sep,
  address      = {Stockholm},
  isbn         = {978-92-9193-311-2},
  doi          = {10.2900/58229},
  url          = {https://www.ecdc.europa.eu/sites/default/files/media/en/publications/Publications/1109_TER_evidence_based_methods_for_public_health.pdf}
}

@misc{WHO2022LivingGuidelines,
  title        = {The Living Approach to WHO normative products and country implementation: Member State Briefing},
  author       = {{World Health Organization}},
  year         = {2022},
  month        = oct,
  day          = {31},
  howpublished = {Member State Briefing (PDF slides)},
  url          = {https://apps.who.int/gb/mspi/pdf_files/2022/10/Item2_31-10.pdf}
}

@misc{WHO2024DisinformationPublicHealth,
  title        = {Disinformation and public health},
  author       = {{World Health Organization}},
  year         = {2024},
  month        = feb,
  day          = {6},
  howpublished = {Questions and answers},
  url          = {https://www.who.int/news-room/questions-and-answers/item/disinformation-and-public-health},
  note         = {Accessed 2026-02-06}
}

@article{doNascimento2022InfodemicsReviewOfReviews,
  title        = {Infodemics and health misinformation: a systematic review of reviews},
  author       = {Borges do Nascimento, Israel J{\'u}nior and Pizarro, Ana Beatriz and Almeida, Jussara M. and Azzopardi-Muscat, Natasha and Gon{\c{c}}alves, Marisa A. and Bj{\"o}rklund, Mattias and Novillo-Ortiz, David},
  journal      = {Bulletin of the World Health Organization},
  year         = {2022},
  month        = sep,
  day          = {1},
  volume       = {100},
  number       = {9},
  pages        = {544--561},
  doi          = {10.2471/BLT.21.287654},
  pmid         = {36062247},
  pmcid        = {PMC9421549},
  url          = {https://pmc.ncbi.nlm.nih.gov/articles/PMC9421549/}
}
\endgroup
\end{document}